\documentclass[10pt,twocolumn,letterpaper]{article}

\usepackage[pagenumbers]{wacv} 

\definecolor{wacvblue}{rgb}{0.21,0.49,0.74}
\usepackage[pagebackref,breaklinks,colorlinks,allcolors=wacvblue]{hyperref}

\usepackage{pifont}
\usepackage{multirow}
\usepackage{tabularx}
\usepackage{tikz}

\newcommand{\circnum}[1]{%
\tikz[baseline=(char.base)]{
    \node[shape=circle,draw,inner sep=1pt] (char) {#1};
}}

\def\wacvPaperID{1150}
\def\confName{WACV}
\def\confYear{2027}

\title{ReText: Text Boosts Generalization in Image-Based Person Re-identification}

\author{
Karina Kvanchiani$^{1*}$ \\
{\tt\small karinakvanciani@gmail.com}
\and
Timur Mamedov$^{1,2*}$ \\
{\tt\small me@timmzak.com}
Timur Mamedov$^{1,2*}$ \\
{\tt\small me@timmzak.com}
Timur Mamedov$^{1,2*}$ \\
{\tt\small me@timmzak.com}
\and
$^1$Tevian, Russia \quad $^2$Lomonosov Moscow State University, Russia}

\author{
    Timur Mamedov$^{1,2*}$ \quad Karina Kvanchiani$^{1*}$ \quad Anton Konushin$^{2}$ \quad Vadim Konushin$^{1}$ \\
    $^1$Tevian, Moscow, Russia \quad $^2$Lomonosov Moscow State University \\
    {\tt\small me@timmzak.com} \quad {\tt\small karinakvanciani@gmail.com} \quad {\tt\small konushinas@my.msu.ru} \quad {\tt\small vadim@tevian.ai}
}

\begin{document}
\maketitle
\renewcommand{\thefootnote}{}
\footnotetext{$^*$Both authors contributed equally.}
\renewcommand{\thefootnote}{\arabic{footnote}}
\begin{abstract}
Generalizable image-based person re-identification (Re-ID)~aims to recognize individuals across cameras in unseen domains without retraining. While visual features are sensitive to domain shifts, natural language descriptions remain consistent across domains, regardless of lighting changes and other biases. Since well-known multi-camera datasets lack textual descriptions, we update them with captions through a fine-tuned Vision-Language Model~(VLM). We propose ReText, a novel method trained with textual supervision to enrich visual features with semantic cues. Besides, inspired by recent findings showing that incorporating single-camera data into training improves generalization, we also adopt this strategy. However, in contrast to such prior works, we exploit not only visual information but also the textual annotations naturally accompanying these samples. Finally, during training, ReText jointly optimizes three tasks: (1)~Re-ID, (2)~image-text matching, and (3) image reconstruction guided by text. Experiments demonstrate that ReText achieves strong generalization and significantly outperforms state-of-the-art methods on cross-domain Re-ID benchmarks. The expanded multi-camera datasets and VLM weights are publicly available at \href{https://github.com/timmzak/ReText}{https://github.com/timmzak/ReText}.
\end{abstract}
    
\section{Introduction}
\label{sec:intro}

Image-based person re-identification (Re-ID) is the task of recognizing individuals across non-overlapping cameras. While modern Re-ID methods achieve strong performance in single-domain scenarios, their effectiveness often drops significantly when applied to unseen domains without retraining. This setting, known as generalizable person Re-ID, is essential for real-world applications where collecting labeled data for each new environment is infeasible.

Achieving generalization in Re-ID is particularly challenging due to the lack of large-scale and diverse multi-camera data. Such datasets are expensive and labor-intensive to collect, as they require synchronized camera networks and identity annotations across views. Thus, Re-ID models overfit to the training domain and fail to generalize to new domains with other visual and view distributions.

Recent works attempt to address the domain gap by designing increasingly complex architectures~\cite{liao2020interpretable, liao2021transmatcher} and training strategies \cite{ni2023part, cho2024generalizable}. Another line of research explores single-camera data, which is stylistically diverse and easy to collect. ReMix \cite{mamedov2025remix} and DynaMix \cite{mamedov2025dynamix} show that combining multi-camera and single-camera data improves generalization but does not fully exploit the semantic potential of the latter. Next, we show that the lack of the cross-view variation in single-camera data limits its effectiveness without additional supervision~(\cref{sec:proof-of-concept}).

\begin{figure}[t]
    \centering
    \includegraphics[width=0.8\linewidth]{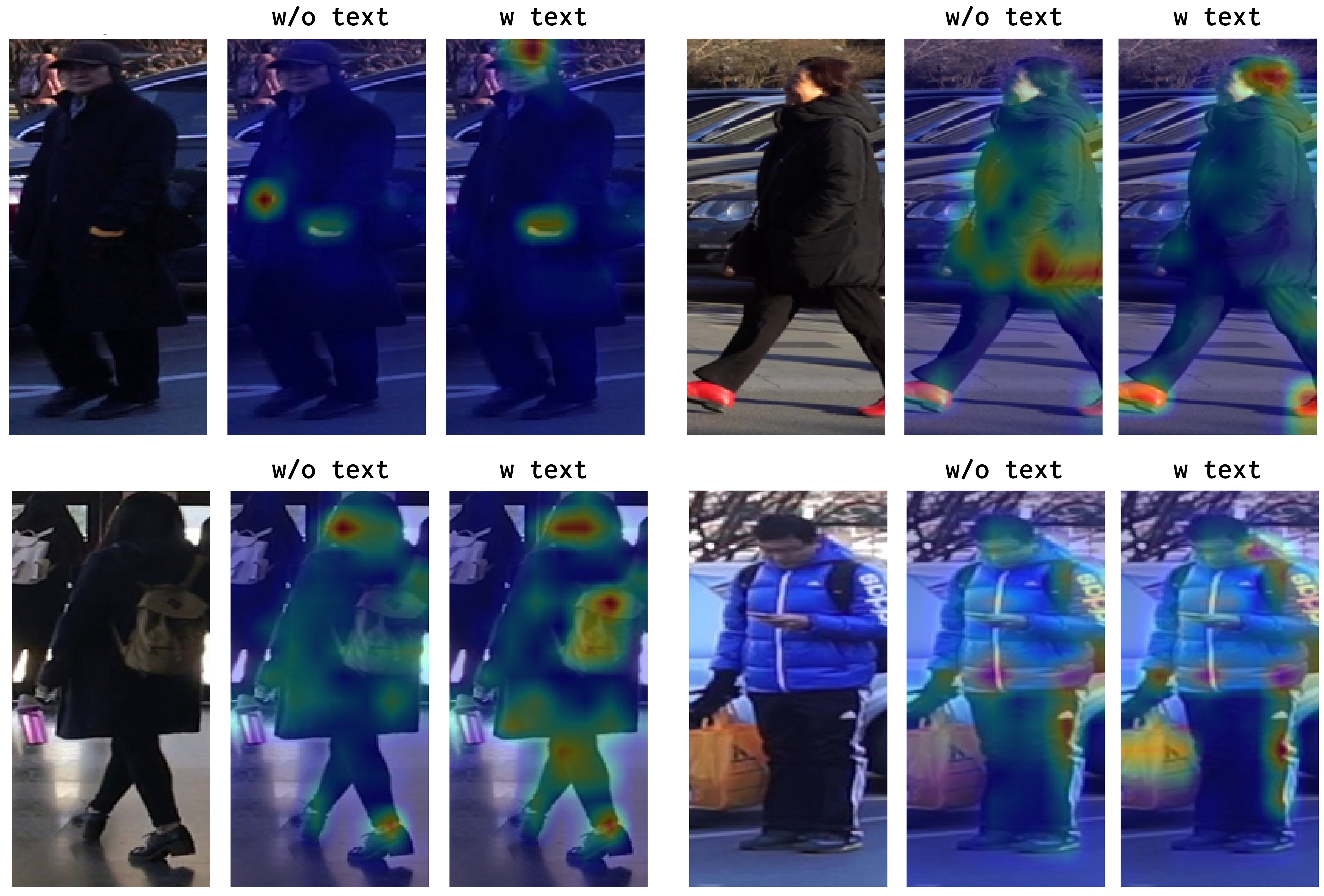}
    \caption{Visual comparison of attention maps for models trained with and without textual supervision. The text-supervised model tends to focus on discriminative attributes like clothing and accessories instead of spreading attention over less informative areas. Attention maps are obtained from the last-layer [CLS]-to-patch attention, averaged across heads, and min-max normalized.}
    \label{fig:attn}
\end{figure}

In parallel, methods like CLIP-ReID \cite{li2023clip} investigate textual supervision for person Re-ID by introducing language as an additional modality. However, they do not exploit single-camera data, and multi-camera datasets lack descriptive captions, so these methods typically rely on learnable text tokens. Thus, the semantic richness and generalization capability of natural language remain underutilized.

We argue for using complete descriptive captions instead of learnable text tokens to fully leverage the advantages of the text modality. Our key insight is that natural language is inherently more domain-invariant than images, as it preserves high-level appearance semantics regardless of domain-specific factors such as lighting, camera properties, and dataset biases (e.g., a pink jacket remains described as pink). Therefore, we supply commonly used multi-camera datasets with text via a fine-tuned Vision-Language Model~(VLM) and show that adding text helps the model focus on distinctive human attributes (\cref{fig:attn}) and yields significant gains. Furthermore, inspired by ReMix~\cite{mamedov2025remix}, we complement our pipeline with single-camera data to improve generalization due to its stylistic diversity and ease of collection. Unlike the prior study, we additionally leverage descriptive captions to provide semantic supervision and reinforce domain-invariant cues, yielding additional gains.

The proposed method, 
ReText, jointly optimizes three tasks during training: (1)~Re-ID, (2) image-text matching (ITM), and (3) image reconstruction (IR) guided by text. The first task drives cross-view person matching, building robustness to viewpoint and appearance changes across cameras. The last two tasks promote generalization by grounding visual representations in domain-invariant semantic cues. 
Our 
contributions are summarized as follows:
\begin{itemize}
    \item We present ReText, a novel method that jointly leverages multi-camera Re-ID data and single-camera data, both with textual descriptions.
    \item We design a three-task training strategy: Re-ID for identity-discriminative learning, ITM for cross-modal alignment, and IR guided by text to enhance robustness to missing or occluded visual information.
    \item We introduce Identity-aware Matching and Structure-preserving losses that explicitly address the specifics of Re-ID, ensuring both cross-modal alignment and intra-class consistency for effective ITM.
    \item We fine-tune VLM and generate textual descriptions with different granularities for five commonly used multi-camera datasets. The fine-tuned VLM weights and expanded datasets are publicly available to promote further research into methods supporting multimodal queries.
\end{itemize}

\vspace{2pt}\noindent Our extensive experiments demonstrate that ReText significantly outperforms state-of-the-art methods across all cross-domain Re-ID benchmarks.
\section{Related Work}
\label{sec:related}

\subsection{Generalizable Person Re-ID}

Generalizable person Re-ID focuses on training models that perform well in unseen domains without retraining. Prior works in this area have explored various strategies to improve cross-domain generalization, including architecture-level modifications \cite{liao2020interpretable, liao2021transmatcher}, normalization techniques \cite{jiao2022dynamically, qi2022novel, jin2023style}, and meta-learning approaches~\cite{choi2021meta, zhao2021learning}. Other methods have focused on domain-invariant feature learning~\cite{zhang2022adaptive, lin2023deep} and part-aware modeling \cite{ni2023part}. In addition to architectural and training innovations, more effective mini-batch sampling approaches have been proposed to encourage domain robustness during training \cite{liao2022graph, zhao2024clip}.

\begin{figure*}[t]
    \centering
    \includegraphics[width=0.8\linewidth]{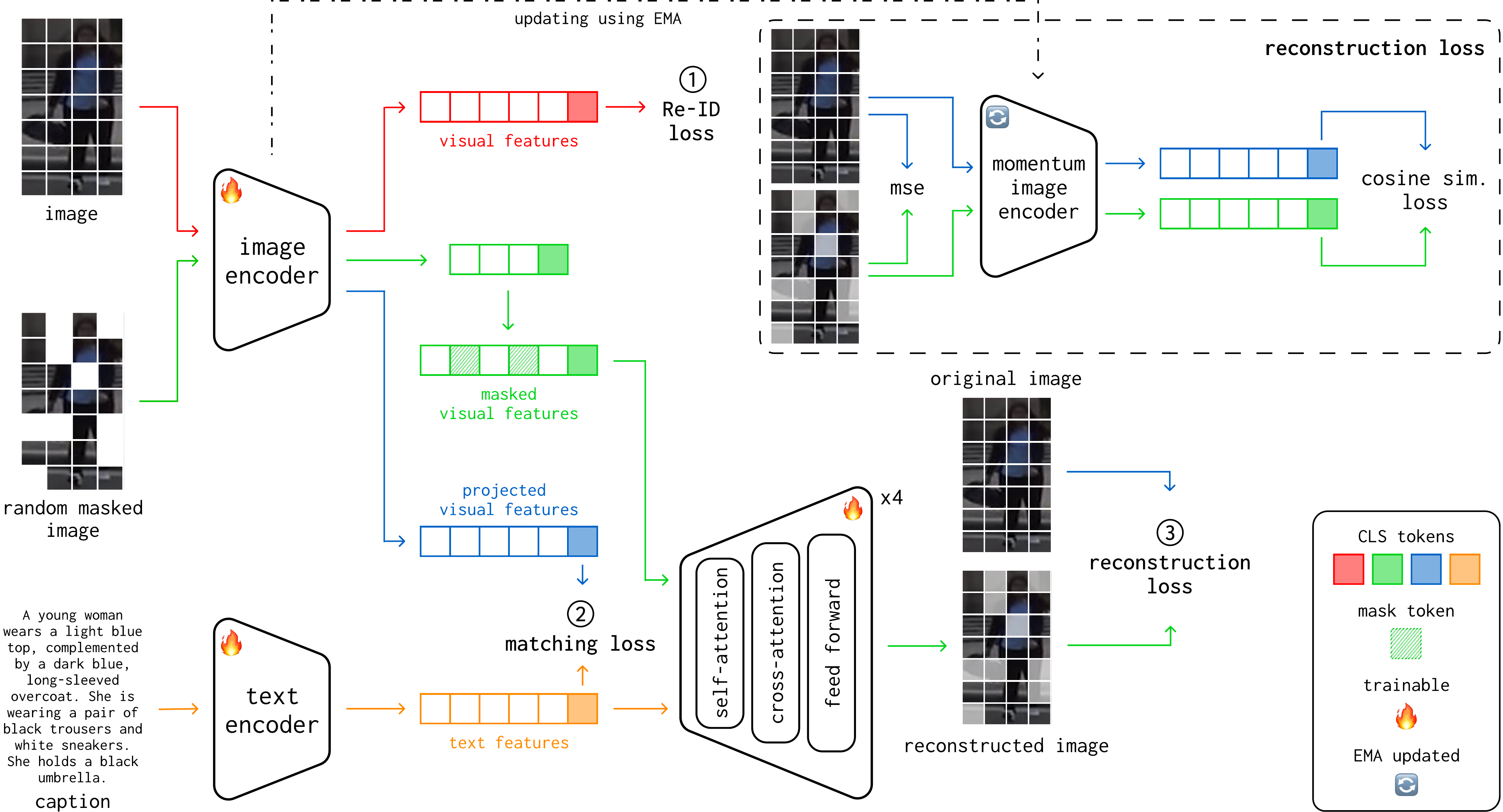}
    \caption{Overview of ReText. The model receives three types of inputs: (a) images from multi-camera, single-camera, or both data sources, (b) masked versions of images, and (c) textual captions. It jointly optimizes the Re-ID loss, the image-text matching loss, and the text-guided reconstruction loss.}
    \label{fig:method-scheme}
\end{figure*}

Despite significant progress in model architectures and training strategies, the role of training data diversity and language-based supervision remains underexplored in generalizable person Re-ID. ReText highlights the importance of both: we demonstrate that joint training on a mixture of multi-camera Re-ID data and stylistically diverse single-camera data with textual descriptions significantly boosts domain generalization and achieves state-of-the-art performance across multiple cross-domain Re-ID benchmarks.

\subsection{Use of Single-Camera Data}

Single-camera data offers a promising source of training samples due to its abundance and stylistic diversity. It can be collected at scale through automated processing of publicly available video streams, such as YouTube videos~\cite{fu2021unsupervised, fu2022large}. However, from the perspective of the person Re-ID task, single-camera data is inherently simpler, as it lacks cross-view variation --- a core challenge in Re-ID. Consequently, it has primarily been used for self-supervised pre-training, followed by fine-tuning on multi-camera Re-ID datasets~\cite{mamedov2023approaches, hu2024personvit}. In contrast, ReMix \cite{mamedov2025remix} and DynaMix~\cite{mamedov2025dynamix} proposed a joint training procedure that combines both multi-camera and single-camera data. This approach significantly improved generalization, showing the value of single-camera data when incorporated correctly.

Yet, ReMix and DynaMix overlook the semantic potential of single-camera data, while ReText leverages it as both a source of stylistic diversity and language-supervised training. Combining such data with descriptive captions introduces semantic complexity, which enhances generalization.

\subsection{Multimodal Learning}

Multimodal models such as CLIP \cite{radford2021learning}, ALBEF \cite{li2021align}, and BLIP \cite{li2022blip} have demonstrated strong cross-modal learning capabilities. Building on these foundations, recent works have introduced multimodal learning into person Re-ID \cite{li2023clip, zhao2024clip, zhao2025cilp}. Such methods optimize learnable text tokens to encode identity-specific semantics and fine-tune the image encoder using these tokens through the CLIP loss. 

Since the CLIP loss treats images of the same identity as different classes, it is suboptimal for person Re-ID, where such images are often similar. Our proposed Identity-aware Matching and Structure-preserving losses overcome this drawback, while the Reconstruction Loss 
improves robustness to missing or occluded visual information.

\subsection{Unified Person Re-ID Methods}

Existing unified Re-ID methods, such as Instruct-ReID~\cite{he2024instruct}, Instruct-ReID++~\cite{he2025instruct}, and HPL~\cite{zhou2025hierarchical} aim to handle multiple Re-ID scenarios within a single model: image-based, text-based, visible-infrared, clothes-changing, etc. These works neither use text to increase the complexity of single-camera data nor study how this affects cross-dataset generalization.

In contrast, ReText focuses solely on image-based person Re-ID and is tailored specifically to improve visual representations by anchoring them to texts.
\section{Proposed Method}
\label{sec:method}

ReText consists of four core components: an image encoder, a momentum image encoder, a text encoder, and an image decoder (\cref{fig:method-scheme}). During inference, only the momentum image encoder is used, meaning the inference is image-only. 
Such image encoder modification is widely used in recent person Re-ID methods \cite{mamedov2025remix, mamedov2025dynamix}, which updates via exponential moving average and serves to stabilize training.

During training, ReText receives images from multi-camera data, single-camera data, or both, their masked versions, and corresponding textual captions. The model jointly optimizes the following three tasks: (1)~Re-ID task~(\cref{sec:re-id-task}), (2)~image-text matching (ITM) task~(\cref{sec:itm-task}), and (3)~image reconstruction (IR) task on masked images guided by texts (\cref{sec:reconstruction-task}). 
All three tasks are complementary, allowing ReText to learn discriminative, semantically grounded, and domain-invariant representations.

\subsection{Formal Definitions}

Let $\{(x_i, t_i, y_i)\}_{i=1}^N$ denote a batch of images $x_i$, their corresponding textual captions $t_i$, and identity labels $y_i$. The image encoder $f_\theta$ and text encoder $g_\phi$ are both transformer-based. For both modalities, the $[CLS]$ token serves as a global representation, followed by $W_{img}$ and $W_{txt}$ linear projection heads that map the modality-specific embeddings into a shared embedding space:
\begin{equation}
    z_i^{img} = W_{img} CLS(f_\theta(x_i)),\: z_i^{txt} = W_{txt} CLS(g_\phi(t_i)).
\end{equation}
Hereafter, $\hat{z}_i^{img}$ and $\hat{z}_i^{txt}$ denote the $\ell_2$-normalized versions of $z_i^{img}$ and $z_i^{txt}$, respectively.

\subsection{Re-ID Task}
\label{sec:re-id-task}

The Re-ID task focuses on learning identity-discriminative features across different camera views. To this end, ReText adopts the loss formulation from ReMix and DynaMix, which combines four components: the Instance ($\mathcal{L}_{ins}$), Augmentation ($\mathcal{L}_{aug}$), Centroids~($\mathcal{L}_{cen}$), and Camera Centroids ($\mathcal{L}_{cc}$) losses. The total Re-ID loss is defined as:
\begin{equation}
    \mathcal{L}_{ReID} = \mathcal{L}_{ins} + \mathcal{L}_{aug} + \mathcal{L}_{cen} + 0.5 \cdot \mathcal{L}_{cc}.
    \label{eq:full-loss}
\end{equation}
The Instance Loss pulls anchors closer to positive instances while distancing them from negative ones, promoting a more generalized model solution. The Augmentation Loss aligns augmented images with their originals and separates them from other identities, addressing inter-instance similarity changes from augmentations. The Centroids Loss brings instances closer to their corresponding centroids and pushes them away from other centroids. The Camera Centroids Loss groups instances with the same label but captured by different cameras around their respective centroids.

\subsection{Image-text Matching Task}
\label{sec:itm-task}

To exploit the semantic information provided by natural language, ReText introduces the image-text matching task. This task encourages the model to align visual and textual modalities, thereby learning semantically meaningful and domain-invariant representations.

For this task, we propose two new loss functions that explicitly address the specifics of Re-ID. The Identity-aware Matching Loss ($\mathcal{L}_{im}$), unlike one-to-one matching in CLIP, learns a distribution over multiple positive image-text pairs, enabling identity-aware alignment. The Structure-preserving Loss ($\mathcal{L}_{sp}$) enhances the structure of image representations in the shared embedding space. These losses are jointly optimized to encourage cross-modal alignment while preserving the intra-class structure among image embeddings. The total 
ITM loss is defined as:
\begin{equation}
    \mathcal{L}_{match} = \mathcal{L}_{im} + \mathcal{L}_{sp}.
\end{equation}

\vspace{2pt}\noindent\textbf{The Identity-aware Matching Loss.} To better capture identity-aware alignment between images and texts, ReText formulates the matching objective as a divergence between the predicted similarity distribution and a soft target distribution over all positive image-text pairs.

Given a projected image embedding $z_i^{img}$ and a set of normalized projected text embeddings $\{\hat{z}_j^{txt} \}_{j=1}^N$, the similarity-based probability distribution is computed as:
\begin{equation}
    p_{i,j} = \frac{\exp\left((z_i^{img})^\top \hat{z}_j^{txt}\right)}{\sum_{k=1}^N \exp\left((z_i^{img})^\top \hat{z}_k^{txt}\right)}.
\end{equation}
The corresponding soft target distribution $q_{i,j}$ is defined based on identity labels:
\begin{equation}
    q_{i,j} =
        \begin{cases}
        \alpha, & \text{if } i = j \\
        \frac{1 - \alpha}{N_{pos}}, & \text{if } y_i = y_j \text{ and } i \ne j \\
        0, & \text{otherwise}
    \end{cases},
    \label{eq:q-in-im-loss}
\end{equation}
where $N_{pos}$ is the number of positive instances in the mini-batch with identity $y_i$, and $\alpha \in (0, 1)$ controls the relative weight of the main image-text pair versus other positives.

The image-to-text component of the loss is defined as a KL-divergence between these two distributions:
\begin{equation}
    \mathcal{L}_{it} = \frac{1}{N} \sum_{i=1}^{N} \sum_{j=1}^{N} p_{i,j} \log \frac{p_{i,j}}{q_{i,j} + \epsilon},
\end{equation}
where $\epsilon$ is a small constant for numerical stability.

The text-to-image component $\mathcal{L}_{ti}$ is computed in the same way, by swapping the roles of image and text embeddings. So, the final Identity-aware Matching Loss combines both directions:
\begin{equation}
    \mathcal{L}_{im} = \mathcal{L}_{it} + \mathcal{L}_{ti}.
\end{equation}
Compared to CLIP-style contrastive loss, which performs strict one-to-one matching, our loss learns a distribution over multiple positive pairs. It assigns relative weight to both the main image-text pair and other same-identity instances, enabling more flexible and semantically grounded supervision. This structure-aware formulation improves generalization by capturing richer intra-class variation~(\cref{sec:ablation-matching-loss}).

\vspace{2pt}\noindent\textbf{The Structure-preserving Loss.} While the Identity-aware Matching Loss aligns image and text embeddings across modalities, it does not explicitly enforce structural consistency among image embeddings of the same identity. To address this, we introduce the Structure-preserving Loss, which encourages the preservation of intra-class structure within the shared embedding space. The loss focuses on the hardest positive instance for each anchor and is defined as:
\begin{equation}
    \mathcal{L}_{sp} = \mathbb{E} \left[ -\log \frac{s_{ij}}{s_{ij} + \sum_{k=1}^{K} \exp\left(\langle \hat{z}_i^{img} \cdot \hat{z}_k^{img}\rangle / \tau \right)} \right],
    \label{eq:sp-loss}
\end{equation}
where $s_{ij} = \exp\left(\langle \hat{z}_i^{img} \cdot \hat{z}_j^{img}\rangle / \tau \right)$ denotes the exponential of the temperature-scaled similarity between anchor $i$ and instance $j$, $j = \arg\min_{j=1,..,N_{pos}}\left(\langle \hat{z}_i^{img} \cdot \hat{z}_j^{img}\rangle \right)$ is the index of the hardest positive pair for anchor $i$, $K = N - N_{pos} - 1$ is the count of negative instances not associated with identity $y_i$ in the mini-batch, and $\tau$ is a temperature hyperparameter.

\subsection{Image Reconstruction Task}
\label{sec:reconstruction-task}

To further align visual and textual modalities and enrich semantic understanding, ReText introduces the text-guided image reconstruction task. The objective is to reconstruct the masked patches of an input image by leveraging both the visible visual tokens and the corresponding textual description. This task reinforces the ability of the model to ground semantic cues in visual content and enhances its robustness to missing or occluded information.

Given an image $x_i$, 
randomly selected patches are removed and a masked image $\tilde{x}_i$ is obtained. The visible~(unmasked) patches are passed through the image encoder $f_\theta$, while the masked positions are replaced with learnable mask tokens. In parallel, the corresponding textual caption $t_i$ is encoded via the text encoder $g_\phi$. Both representations are then fed into the decoder $d_\psi$, which predicts the pixel values for the missing patches.

\vspace{2pt}\noindent\textbf{The Reconstruction Loss.} ReText employs two complementary loss functions to supervise the image reconstruction process. The MSE loss encourages accurate recovery of the original image patches:
\begin{equation}
    \mathcal{L}_{mse} = \frac{1}{M} \sum_{m=1}^{M} \left\| x_i^{(m)} - \hat{x}_i^{(m)} \right\|_2^2,
\end{equation}
where $M$ is the number of masked patches, $x_i^{(m)}$ and $\hat{x}_i^{(m)}$ are the ground truth and predicted patches, respectively.

To complement this, the Cosine Similarity loss ensures that the reconstructed image retains the original's high-level semantics. Specifically, both the original $x_i$ and reconstructed $\hat{x}_i$ images are passed through the frozen momentum image encoder $f{_\theta’}$, and their resulting $[CLS]$ tokens are aligned via cosine similarity:
\begin{equation}
    \mathcal{L}_{cos} = 1 - \cos\left( CLS(f_{\theta’}(x_i)), CLS(f_{\theta’}(\hat{x}_i)) \right).
\end{equation}
The Reconstruction Loss combines both components:
\begin{equation}
    \mathcal{L}_{rec} = \mathcal{L}_{mse} + \mathcal{L}_{cos}.
\end{equation}
This dual loss design encourages not only pixel-level correctness but also the preservation of identity-specific semantics, critical for robust cross-domain generalization.

While standard Re-ID objectives in ReText handle viewpoint invariance, the adapted Reconstruction Loss, on par with the The Identity-aware Matching Loss, promotes semantic consistency between visual and textual features, adding more domain-agnostic information from texts.
\section{Expanding Image-Based Re-ID Datasets}
\label{sec:texts_for_i2i}

\begin{table}[tb]
    \centering
    \scalebox{0.76}{
    \begin{tabular*}{0.55\textwidth}{@{\hspace{3pt}}@{\extracolsep{\fill}}lcccccc}
        \toprule
        Dataset & Multi & Text & \#Images & \#IDs & \#Cams \\
        \midrule
        CUHK03-NP & \ding{51} & \ding{55} & $14,097$ & $1,467$ & $2$ \\
        Market-1501 & \ding{51} & \ding{55} & $32,217$ & $1,501$ & $6$ \\
        MSMT17 & \ding{51} & \ding{55} & $126,441$ & $4,101$ & $15$ \\
        CUHK-SYSU & \ding{51} & \ding{55} & $34,574$ & $11,934$ & --- \\
        RandPerson & \ding{51} & \ding{55} & $132,145$ & $8,000$ & $19$ \\
        \midrule
        SYNTH-PEDES & \ding{55} & \ding{51} & $4,791,771$ & $312,321$ & --- \\
        \bottomrule
    \end{tabular*}}
    \caption{Statistics of datasets used in our experiments. The single-camera SYNTH-PEDES is significantly larger than all multi-camera datasets and textually annotated.}
    \label{tab:datasets}
\end{table}

\begin{figure}[t]
    \centering
    \includegraphics[width=0.8\linewidth]{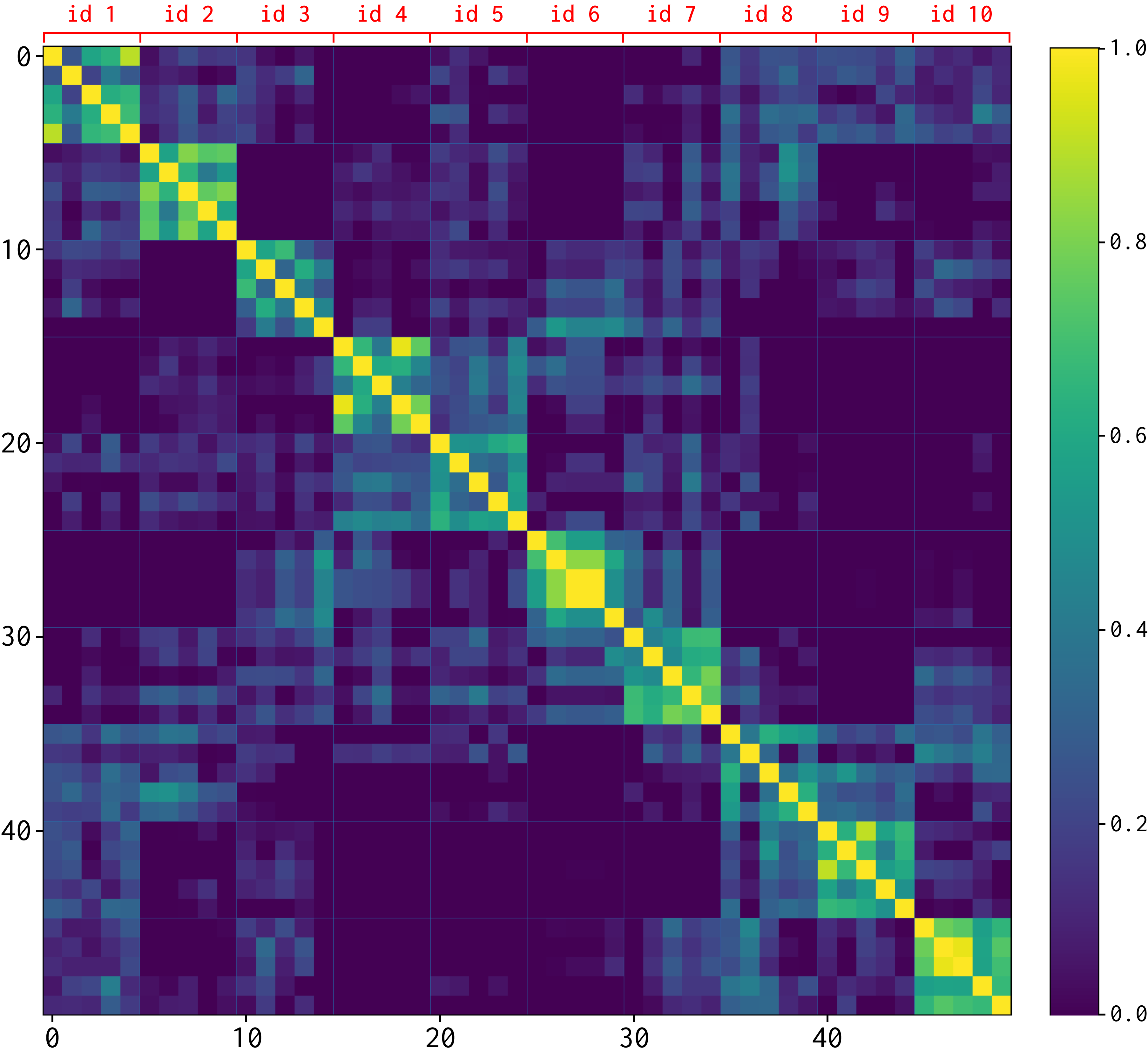}
    \caption{The block-structured similarity matrix. The caption embeddings are obtained from the text encoder trained within ReText on MSMT17 with textual supervision.}
    \label{fig:sim}
\end{figure}

To assess the impact of text supervision as a multi-camera domain-invariant anchor, we generate texts for the first 5 datasets from \cref{tab:datasets}.

\vspace{2pt}\noindent\textbf{VLM Fine-tuning.} We employ the Vision-Language Model (VLM) to generate texts for each image. VLM is preliminary fine-tuned on the UFine3C \cite{zuo2024ufinebench} benchmark, containing reorganized samples from the CUHK-PEDES~\cite{li2017person}, ICFG-PEDES \cite{ding2021semantically}, and UFine6926 \cite{zuo2024ufinebench} datasets to train the model to produce coarse-grained, medium-grained, and fine-grained texts, respectively. The prompt instruction is the same for fine-tuning and captioning stages: ``Describe the person focusing only on stable visual features across cameras, including clothing, shoes, hairstyle, accessories, and carried objects, without mentioning body shape, height, face, emotions, pose, actions, background, or context.'' More details of the captioning process and its efficiency are included in the supplementary material~(\cref{sec:data-for-llava}).

\vspace{2pt}\noindent\textbf{Caption Filtering.} Despite the overall reliability of the fine-tuned VLM, generated captions occasionally exhibit repetition loops, where autoregressive generation produces repeated phrases or sentences. To address this, we apply a two-stage filtering strategy: (1) repetition-based filtering detects and deduplicates repeated n-gram subsequences, retaining only the first occurrence of each repeated phrase; (2) length-based filtering discards captions exceeding the empirically determined limit of 150 words, as excessively long outputs typically indicate 
off-prompt content. Qualitative examples of these failure modes, along with a manual attribute-level verification of caption quality, 
are provided in \cref{sec:quality-of-caps} 
of the supplementary material. The fine-tuned VLM naturally produces descriptions of varying granularity, 
reflecting the visual complexity of each image~(see \cref{sec:stat-of-caps} in the supplementary material for details).


\vspace{2pt}\noindent\textbf{Intra-Identity Caption Consistency.} To analyze the intra-identity consistency of the generated captions, we construct a block-structured similarity matrix over caption embeddings. 
Specifically, we randomly select 10 identities, each associated with 5 image-text pairs, and compute pairwise cosine similarities between all caption embeddings, ordered by identity, to assess intra- and inter-identity semantic consistency.
As 
shown in \cref{fig:sim}, distinct bright blocks along the main diagonal, corresponding to captions of the same identity, form clear high-similarity patterns, indicating strong intra-identity semantic consistency. In contrast, off-diagonal regions, representing inter-identity comparisons, exhibit significantly lower similarity values, confirming that the generated captions are discriminative across different identities. 
The analysis demonstrates that the generated texts capture identity-specific characteristics of multi-camera data.
\section{Experiments}
\label{sec:experiments}

\subsection{Datasets and Evaluation Metrics}

\vspace{2pt}\noindent\textbf{Multi-camera Datasets.} We evaluate the proposed ReText method on well-known multi-camera datasets: CUHK03-NP \cite{li2014deepreid}, Market-1501 \cite{zheng2015scalable}, MSMT17~\cite{wei2018person}, CUHK-SYSU\footnote{Although each identity in CUHK-SYSU appears under a single camera view, standard evaluation protocols still treat it as multi-camera data.}~\cite{xiao2017joint}, and a subset of RandPerson \cite{wang2020surpassing} (\cref{tab:datasets}). Although each mentioned dataset is expanded with generated texts as described in \cref{sec:texts_for_i2i}, we provide their initial statistics in the \cref{tab:datasets} for clarity. We do not include the multi-camera dataset DukeMTMC-reID \cite{ristani2016performance} in our experiments, as it was withdrawn due to ethical concerns.

\vspace{2pt}\noindent\textbf{Single-camera Dataset.} We use SYNTH-PEDES \cite{zuo2024plip} as the single-camera dataset with textual annotations. It is constructed from images sourced from LUPerson-NL \cite{fu2022large} and LPW \cite{song2018region}, and enriched with automatically generated captions using a dedicated annotation framework. As seen in \cref{tab:datasets}, SYNTH-PEDES is significantly larger than traditional multi-camera Re-ID datasets and offers much broader diversity. It is important to note that this single-camera dataset does not overlap with any of the multi-camera datasets used in our experiments, which prevents data leakage and ensures a fair evaluation.

\vspace{2pt}\noindent\textbf{Metrics.} In our experiments, we use Cumulative Matching Characteristics~($Rank_1$) and mean Average Precision~($mAP$) as evaluation metrics.

\subsection{Implementation Details}
\vspace{2pt}\noindent\textbf{Re-ID Model Training.} Both image encoder and momentum image encoder are ViT-Base \cite{dosovitskiy2020image}, pre-trained in a self-supervised manner on the LUPerson dataset \cite{fu2021unsupervised}. The text encoder is BERT-Base \cite{devlin2019bert}. We use the AdamW optimizer with a learning rate of $10^{-5}$ and a weight decay rate of $0.02$. A warm-up scheme is applied during the first $10$ epochs. 
Training is stopped early if the loss on the held-out part of the training set does not improve for $5$ consecutive epochs. ReText is trained for up to $100$ epochs until convergence on 2 NVIDIA GeForce RTX 3090 GPUs with a mini-batch size of $96$, comprising $32$ images from multi-camera data and $64$ from single-camera data. All images are resized to $256 \times 128$ and augmented with random cropping, horizontal flipping, Gaussian blurring, and random grayscale transformations.
More details about mini-batch composition, decoder architecture, training time, and parameter analysis are provided in \cref{sec:additional-implementation-details} and \cref{sec:parameter-analysis} of the supplementary material.

\vspace{2pt}\noindent\textbf{VLM Fine-tuning.} We use LLaVA-1.6-7B as the VLM to generate textual descriptions for the multi-camera datasets. We fine-tune it for $2$ epochs with the AdamW optimizer at a learning rate of $2^{-5}$ on 6 NVIDIA RTX A5000 GPUs.

\subsection{Proof-of-Concept}
\label{sec:proof-of-concept}

\begin{table}[tb]
    \centering
    \fontsize{7.5pt}{11pt}\selectfont
    \tabcolsep=2.5pt
    \scalebox{0.8}{
    \begin{tabular*}{0.6\textwidth}{@{\extracolsep{\fill}}c| c@{\hskip 7pt}c@{\hskip 7pt}c@{\hskip 7pt}c|cc|cc|cc}
        \toprule
        \multirow{2}{*}{\#} & Multi & Multi & Single & Single & \multicolumn{2}{c|}{CUHK03-NP} & \multicolumn{2}{c|}{Market-1501} & \multicolumn{2}{c}{Average} \\
        & w/o text & w. text & w/o text & w. text & $Rank_1$ & $mAP$ & $Rank_1$ & $mAP$ & $Rank_1$ & $mAP$ \\
        \midrule
        \circnum{1} & \ding{51} & & & & $42.8$ & $43.3$ & $89.1$ & $71.7$ & $66.0$ & $57.5$ \\
        \circnum{2} & & \ding{51} & & & $\underline{61.9}$ & $61.8$ & $\underline{93.3}$ & $81.4$ & $\underline{77.6}$ & $71.6$ \\
        \circnum{3} & & & \ding{51} & & $24.8$ & $26.9$ & $84.1$ & $66.3$ & $54.5$ & $46.6$ \\
        \circnum{4} & & & & \ding{51} & $36.7$ & $36.8$ & $86.4$ & $68.5$ & $61.6$ & $52.7$ \\
        \circnum{5} & \ding{51} & & \ding{51} & & $49.9$ & $51.3$ & $92.4$ & $78.2$ & $71.2$ & $64.8$ \\
        \circnum{6} & & \ding{51} & & \ding{51} & $\underline{61.9}$ & $\underline{62.9}$ & $93.2$ & $\underline{82.8}$ & $\underline{77.6}$ & $\underline{72.9}$ \\
        \circnum{7} & \ding{51} & & & \ding{51} & $\mathbf{63.4}$ & $\mathbf{63.1}$ & $\mathbf{93.6}$ & $\mathbf{83.6}$ & $\mathbf{78.5}$ & $\mathbf{73.4}$ \\
        \bottomrule
    \end{tabular*}}
    \caption{Effect of single-camera data and textual supervision on cross-domain Re-ID. MSMT17 is used as multi-camera data for training.}
    \label{tab:proof-of-concept-overall}
\end{table}

We validate three key insights behind ReText: (1) supplying multi-camera data with descriptive captions improves the model's generalization ability; (2) while single-camera data is less challenging for person Re-ID due to the lack of cross-view variation, its stylistic diversity can enhance generalization; and (3) combining single-camera data with textual captions provides language supervision that facilitates the learning of domain-invariant representations. The results in \cref{tab:proof-of-concept-overall} confirm all claims.

\vspace{2pt}\noindent\textbf{Textual Supervision Improves Generalization.} Comparing the first two rows isolates the effect of text alone --- adding captions to multi-camera data without any single-camera data yields a $\sim$14\% average $mAP$ gain over the text-free baseline. This demonstrates that natural language provides rich semantic cues~(e.g., clothing, attributes) that help the model learn more 
domain-invariant representations and concentrate on stable semantic attributes (\cref{fig:attn}).

\vspace{2pt}\noindent\textbf{Cross-view Variation is Crucial.} The third row of \cref{tab:proof-of-concept-overall} shows that training solely on single-camera data, without textual supervision, yields significantly lower performance than training on multi-camera Re-ID data (first row). This confirms that single-camera data, lacking cross-view variation, is unsuitable for the Re-ID task, so excluding it from the Re-ID objectives yields the best results (\cref{suppl:single-camera-reid-losses} of the supplementary material). 

\vspace{2pt}\noindent\textbf{Stylistic Diversity Improves Generalization.} The fifth row of \cref{tab:proof-of-concept-overall} shows that joint training on a mixture of multi-camera and single-camera data (without text)\footnote{Note that the described training occurs in ReMix settings, which effectively utilizes both inputs for strong generalization ability.} leads to a clear improvement over using multi-camera data alone~(first row). This indicates that the stylistic diversity present in single-camera data exposes the model to a broader distribution of visual appearance, thereby enhancing its ability to generalize to unseen domains.

\vspace{2pt}\noindent\textbf{Captions And Stylistic Diversity: The Key to Generalization.} Combining stylistic diversity from single-camera data and semantic supervision from captions with multi-camera data (seventh row) leads to the strongest cross-domain performance. However, the sixth row of \cref{tab:proof-of-concept-overall} demonstrates that the expansion of multi-camera and single-camera datasets with texts simultaneously results in a slight drop in quality. We associate this effect with a plateau that occurs as the number of samples with texts increases~(\cref{tab:synth-quant}), concluding that the quantity of texts is more important than the complexity of the images they describe. 
We expect such effect to form the basis for further research.

\vspace{2pt}\noindent\textbf{Conclusions.} 
Our primary goal is to explore how text supervision during training can enhance image-based person Re-ID, and how semantically rich captions can increase the usefulness of single-camera data. Further experiments are provided in two versions: one trained on multi-camera data with texts (Mt)\footnote{ReText (Mt) shows the effect of textual supervision alone, without any additional image data.} and the second on a mixture of multi-camera and single-camera data, with only the latter paired with texts (MSt). The ablation study is based on the MSt version due to its superior results.

\subsection{Comparison with State-of-the-Art Methods}
\label{sec:protocol1}

We compare ReText with other state-of-the-art generalizable person Re-ID methods using a standard evaluation protocol, which implies training on a single multi-camera dataset and testing on a different one. Two other protocols are discussed in \cref{sec:protocols} of the supplementary material to comprehensively demonstrate the advantages of the presented method.

As shown in \cref{tab:comparison-single}, ReText consistently outperforms existing methods across all cross-domain benchmarks. It also surpasses transformer-based models such as TransMatcher~\cite{liao2021transmatcher} and PAT \cite{ni2023part}, confirming that its generalization strength comes from multimodal joint learning on a mixture of multi-camera and single-camera data rather than architectural complexity. All approaches exclude ReMix, DynaMix, and our ReText rely solely on labeled multi-camera datasets, as they are not designed to handle heterogeneous training data. Nonetheless, the ReText~(Mt) surpasses them, even without the additional single-camera data, highlighting the effectiveness of incorporating textual descriptions and the ability of the fine-tuned VLM to generate informative captions. 

The magnitude of improvement varies across target domains --- training on MSMT17, ReText improves DynaMix's mAP by $13.5$ on CUHK03-NP and by $5.9$ on Market-1501. However, normalizing gains by the remaining headroom, ReText closes a comparable proportion of the remaining gap toward the maximum score in both cases.

\begin{table}[t]
    \centering
    \fontsize{8pt}{11pt}\selectfont
    \tabcolsep=2.5pt
    \scalebox{0.77}{
    \begin{tabular}{@{\hspace{7pt}}l|c|cc|cc|cc|cc}
        \toprule
        \multirow{2}{*}{Method} & \multirow{2}{*}{\rotatebox{90}{Tr. D.}} & \multicolumn{2}{c|}{CUHK03-NP} & \multicolumn{2}{c|}{Market-1501} & \multicolumn{2}{c|}{MSMT17} & \multicolumn{2}{c}{Average}\\
        & & $Rank_1$ & $mAP$ & $Rank_1$ & $mAP$ & $Rank_1$ & $mAP$ & $Rank_1$ & $mAP$ \rule[-0.5ex]{0pt}{0pt}\\
        \midrule
        TransMatcher \cite{liao2021transmatcher} & \multirow{9}{*}{\rotatebox{90}{Market-1501}} & $22.2$ & $21.4$ & --- & --- & $47.3$ & $18.4$ & $34.8$ & $19.9$ \rule{0pt}{2.3ex}\\
        QAConv-GS \cite{liao2022graph} & & $19.1$ & $18.1$ & $91.6$ & $75.5$ & $45.9$ & $17.2$ & $32.5$ & $17.6$ \rule{0pt}{2.3ex}\\
        PAT \cite{ni2023part} & & $25.4$ & $26.0$ & $92.4$ & $81.5$ & $42.8$ & $18.2$ & $34.1$ & $22.1$ \rule{0pt}{2.3ex}\\
        LDU \cite{peng2024invariance} & & $18.5$ & $18.2$ & --- & --- & $35.7$ & $13.5$ & $27.1$ & $15.8$ \rule{0pt}{2.3ex}\\
        DCAC \cite{li2025unleashing} & & $33.2$ & $32.5$ & $94.9$ & $86.8$ & $52.1$ & $23.4$ & $42.7$ & $27.9$ \rule{0pt}{2.3ex}\\
        ReMix \cite{mamedov2025remix} & & --- & --- & $96.2$ & $89.8$ & --- & --- & --- & --- \rule{0pt}{2.3ex}\\
        DynaMix \cite{mamedov2025dynamix} & & $51.0$ & $52.2$ & $\mathbf{97.4}$ & $\mathbf{93.8}$ & $65.5$ & $36.7$ & $58.2$ & $44.5$ \rule{0pt}{2.3ex}\\
        ReText (Mt) & & $\mathbf{57.6}$ & $\mathbf{58.2}$ & $\underline{97.2}$ & $\underline{93.4}$ & $\underline{66.2}$ & $\underline{39.4}$ & $\underline{61.9}$ & $\underline{48.8}$ \rule{0pt}{2.1ex}\\
        ReText (MSt) & & $\mathbf{57.6}$ & $\underline{57.5}$ & $\mathbf{97.4}$ & $\mathbf{93.8}$ & $\mathbf{70.1}$ & $\mathbf{43.1}$ & $\mathbf{63.8}$ & $\mathbf{50.3}$ \rule{0pt}{2.1ex}\\
        \midrule
        TransMatcher \cite{liao2021transmatcher} & \multirow{9}{*}{\rotatebox{90}{MSMT17}} & $23.7$ & $22.5$ & $80.1$ & $52.0$ & --- & --- & $51.9$ & $37.2$ \rule{0pt}{2.3ex}\\
        QAConv-GS \cite{liao2022graph} & & $20.9$ & $20.6$ & $79.1$ & $49.5$ & $79.2$ & $50.9$ & $50.0$ & $35.0$ \rule{0pt}{2.3ex}\\
        PAT \cite{ni2023part} & & $24.2$ & $25.1$ & $72.2$ & $47.3$ & $75.9$ & $52.0$ & $48.2$ & $36.2$ \rule{0pt}{2.3ex}\\
        LDU \cite{peng2024invariance} & & $21.3$ & $21.3$ & $74.6$ & $44.8$ & --- & --- & $47.9$ & $33.0$ \rule{0pt}{2.3ex}\\
        DCAC \cite{li2025unleashing} & & $34.4$ & $34.1$ & $77.9$ & $52.1$ & $88.3$ & $70.1$ & $56.2$ & $43.1$ \rule{0pt}{2.3ex}\\
        ReMix \cite{mamedov2025remix} & & $27.3$ & $27.4$ & $78.2$ & $52.4$ & $84.8$ & $63.9$ & $52.8$ & $39.9$ \rule{0pt}{2.3ex}\\
        DynaMix \cite{mamedov2025dynamix} & & $48.9$ & $49.6$ & $91.1$ & $77.7$ & $90.4$ & $76.3$ & $70.0$ & $63.7$ \rule{0pt}{2.3ex}\\
        ReText (Mt) & & $\underline{61.9}$ & $\underline{61.8}$ & $\underline{93.3}$ & $\underline{81.4}$ & $\underline{91.0}$ & $\underline{78.1}$ & $\underline{77.6}$ & $\underline{71.6}$ \rule{0pt}{2.1ex}\\
        ReText (MSt) & & $\mathbf{63.4}$ & $\mathbf{63.1}$ & $\mathbf{93.6}$ & $\mathbf{83.6}$ & $\mathbf{91.4}$ & $\mathbf{78.7}$ & $\mathbf{78.5}$ & $\mathbf{73.3}$ \rule{0pt}{2.1ex}\\
        \midrule
        TransMatcher \cite{liao2021transmatcher} & \multirow{7}{*}{\rotatebox{90}{RandPerson}} & $17.1$ & $16.0$ & $77.3$ & $49.1$ & $48.3$ & $17.7$ & 47.6 & 27.6 \rule{0pt}{2.3ex}\\
        QAConv-GS \cite{liao2022graph} & & $18.4$ & $16.1$ & $76.7$ & $46.7$ & $45.1$ & $15.5$ & 46.7 & 26.1 \rule{0pt}{2.3ex}\\
        PAT \cite{ni2023part} & & $20.2$ & $20.1$ & $73.7$ & $46.9$ & $45.5$ & $19.4$ & 46.5 & 28.8 \rule{0pt}{2.3ex}\\
        ReMix \cite{mamedov2025remix} & & $19.3$ & $18.4$ & $72.7$ & $45.4$ & --- & --- & --- & --- \rule{0pt}{2.3ex}\\
        DynaMix \cite{mamedov2025dynamix} & & $36.9$ & $39.3$ & $89.3$ & $72.4$ & $67.1$ & $36.5$ & 64.4 & 49.4 \rule{0pt}{2.3ex}\\
        ReText (Mt) & & $\underline{44.2}$ & $\underline{44.1}$ & $\underline{90.7}$ & $\underline{76.4}$ & $\underline{67.9}$ & $\underline{39.7}$ & $\underline{67.6}$ & $\underline{53.4}$ \rule{0pt}{2.1ex}\\
        ReText (MSt) & & $\mathbf{46.9}$ & $\mathbf{47.7}$ & $\mathbf{91.7}$ & $\mathbf{79.7}$ & $\mathbf{70.8}$ & $\mathbf{42.3}$ & $\mathbf{69.8}$ & $\mathbf{56.6}$ \rule{0pt}{2.1ex}\\
        \bottomrule
    \end{tabular}}
    \caption{Comparison of ReText with other state-of-the-art methods. Tr. D. denotes the training dataset. Metrics on the Market-1501 and MSMT17 datasets are excluded from average values in the first and second blocks, respectively. 
    }
    \label{tab:comparison-single}
\end{table}

\begin{table}[t]
    \centering
    \fontsize{7pt}{11pt}\selectfont
    \tabcolsep=2.5pt
    \scalebox{0.95}{
    \begin{tabular}{@{\hspace{7pt}}l|cc|cc|cc}
        \toprule
        \multirow{2}{*}{Method}  & \multicolumn{2}{c|}{Market-1501} & \multicolumn{2}{c|}{MSMT17} & \multicolumn{2}{c}{Average}\\
        & $Rank_1$ & $mAP$ & $Rank_1$ & $mAP$ & $Rank_1$ & $mAP$ \rule[-0.5ex]{0pt}{0pt}\\
        \midrule
        \multicolumn{7}{c}{\textbf{Image-Based Re-ID with Text Tokens}} \\
        \midrule
        CLIP-ReID \cite{li2023clip} & $95.4$ & $90.5$ & $89.7$ & $75.8$ & $92.6$ & $83.2$ \rule{0pt}{2.1ex}\\
        CLIP-DFGS \cite{zhao2024clip} & $93.7$ & $85.5$ & $86.4$ & $66.3$ & $90.1$ & $75.9$ \rule{0pt}{2.1ex}\\
        \midrule
        \multicolumn{7}{c}{\textbf{Unified Re-ID Methods}} \\ \midrule
        PLIP \cite{zuo2024plip} & $\underline{97.3}$ & $93.2$ & --- & --- & --- & --- \rule{0pt}{2.3ex}\\
        Instruct-ReID (STL) \cite{he2024instruct} & $96.2$ & $92.3$ & $86.2$ & $71.9$ & $91.2$ & $82.1$ \rule{0pt}{2.3ex}\\
        Instruct-ReID (MTL) \cite{he2024instruct} & $96.5$ & $\underline{93.5}$ & $86.9$ & $72.4$ & $91.7$ & $83.0$ \rule{0pt}{2.3ex}\\
        HPL \cite{zhou2025hierarchical} & $96.0$ & $89.8$ & $\underline{91.0}$ & $\mathbf{79.0}$ & $93.5$ & $84.4$ \rule{0pt}{2.3ex}\\
        \midrule
        \multicolumn{7}{c}{\textbf{Image-Based Re-ID with Texts}} \\ \midrule
        ReText (Mt) & $97.2$ & $93.4$ & $\underline{91.0}$ & $78.1$ & $\underline{94.1}$ & $\underline{85.8}$ \rule{0pt}{2.1ex}\\
        ReText (MSt) & $\mathbf{97.4}$ & $\mathbf{93.8}$ & $\mathbf{91.4}$ & $\underline{78.7}$ & $\mathbf{94.4}$ & $\mathbf{86.3}$ \rule{0pt}{2.1ex}\\
        \bottomrule
    \end{tabular}}
    \caption{Comparison of ReText with other methods that support language supervision through text tokens or by unifying methods for image and text queries. STL and MTL versions of Instruct-ReID mean single-task and multi-task learning, respectively. 
    }
    \label{tab:comparison-single-not-generalize}
\end{table}

Moreover, ReText demonstrates a performance gain compared to learnable text-token approaches (\cref{tab:comparison-single-not-generalize}), indicating their insufficiency. We also compare ReText with the unified method PLIP, which is first pre-trained on the SYNTH-PEDES dataset and then fine-tuned on the target domain. This indicates that the advantages stem from the effectiveness of semantically enriched multimodal supervision rather than from huge additional data. Finally, we compare ReText with other unified approaches such as Instruct-ReID and HPL, motivated by ReText's ability to process two modalities. ReText consistently outperforms them, although the Instruct-ReID MTL version is trained on approximately $5$ million samples across various Re-ID tasks\footnote{Note that since unified methods target a broader multi-task scope, a conclusion about ReText's relative performance is difficult to draw.}. Comparing ReText (Mt) with Instruct-ReID (STL), both trained on a single dataset for a single task, the text supervision in ReText leads to quality improvements.

\subsection{Ablation Study}

\vspace{2pt}\noindent\textbf{Effect of Training Objectives.} We first evaluate the impact of each training task by incrementally adding the proposed objectives. As shown in \cref{tab:ablation-study-overall}, the baseline model trained only with the Re-ID loss (first row) achieves moderate performance\footnote{Although we utilize the ReMix losses, the ReText is based on the ViT-Base architecture rather than ResNet50-IBN~\cite{pan2018two}, explaining their performance difference.}. We utilize Re-ID loss only on the multi-camera data, since incorporating single-camera data into this loss leads to degradation~(\cref{suppl:single-camera-reid-losses} of the supplementary material). When we introduce the image-text matching (ITM) task, performance improves substantially, demonstrating the benefit of leveraging textual supervision from captions. Adding the image reconstruction (IR) task provides further gains. The improvement confirms the robustness of the learned representations, especially under partial visual information and domain shift. Examples of reconstructed images are provided in \cref{sec:image-reconstruction} of the supplementary material.

\begin{table}[t]
\centering
\begin{minipage}[t]{0.47\linewidth}
    \vspace{0pt}
    \centering
    \fontsize{8pt}{11pt}\selectfont
    \tabcolsep=2.5pt
    \scalebox{0.83}{%
    \begin{tabular*}{\linewidth}{@{\hspace{1pt}}@{\extracolsep{\fill}}ccc|cc}
        \toprule
        Re-ID & ITM & IR & $Rank_1$ & $mAP$ \\
        \midrule
        \ding{51} & & & $42.8$ & $43.3$ \\
        \midrule
        \multicolumn{5}{c}{Mt} \\
        \midrule
        \ding{51} & \ding{51} & & $60.3$ & $60.9$ \\
        \ding{51} & \ding{51} & \ding{51} & $61.9$ & $61.8$ \\
        \midrule
        \multicolumn{5}{c}{MSt} \\
        \midrule
        \ding{51} & \ding{51} & & $62.9$ & $62.7$ \\
        \ding{51} & \ding{51} & \ding{51} & $\mathbf{63.4}$ & $\mathbf{63.1}$ \\
        \bottomrule
    \end{tabular*}%
    }
    \caption{Effect of the proposed training objectives. 
    The model is trained on MSMT17 and tested on CUHK03-NP.}
    \label{tab:ablation-study-overall}
\end{minipage}%
\hfill
\begin{minipage}[t]{0.495\linewidth}
    \vspace{0pt}
    \centering
    \fontsize{8pt}{11pt}\selectfont
    \tabcolsep=2.5pt
    \scalebox{0.9}{%
    \begin{tabular*}{\linewidth}{@{\hspace{1pt}}@{\extracolsep{\fill}}l|cc}
        \toprule
        Loss & $Rank_1$ & $mAP$ \\
        \midrule
        CLIP loss & $59.8$ & $60.7$ \\
        Soft CLIP loss & $60.2$ & $61.1$ \\
        $\mathcal{L}_{im}$ (ours) & $62.2$ & $62.3$ \\
        $\mathcal{L}_{im} + \mathcal{L}_{sp}$ (ours) & $\mathbf{62.9}$ & $\mathbf{62.7}$ \\
        \bottomrule
    \end{tabular*}%
    }
    \caption{Effect of different image-text matching losses on cross-domain Re-ID performance. The model is trained on MSMT17 and tested on CUHK03-NP.}
    \label{tab:ablation-study-itm-losses}
\end{minipage}
\end{table}

\begin{table}[t]
    \centering
    \fontsize{8pt}{11pt}\selectfont
    \tabcolsep=2.5pt
    \scalebox{0.83}{
    \begin{tabular}{@{\hspace{5pt}}l|c|cc|cc|cc}
        \toprule
        \multirow{2}{*}{Dataset} & \multirow{2}{*}{Texts} & \multicolumn{2}{c|}{CUHK03-NP} & \multicolumn{2}{c|}{Market-1501} & \multicolumn{2}{c}{Average}\\
        & & $Rank_1$ & $mAP$ & $Rank_1$ & $mAP$ & $Rank_1$ & $mAP$ \rule[-0.5ex]{0pt}{0pt}\\
        \midrule
        MSMT17 & 32,621 & $61.9$ & $61.8$ & $93.3$ & $81.4$ & $77.6$ & $71.6$ \rule{0pt}{2.3ex}\\
        \midrule
        \multirow{5}{*}{SYNTH-PEDES} & 32,621 & $61.4$ & $62.3$ & $93.3$ & $82.5$ & $77.4$ & $72.4$ \rule{0pt}{2.3ex}\\
        & 100,000 & $61.5$ & $62.6$ & $92.7$ & $82.1$ & $77.1$ & $72.4$ \rule{0pt}{2.3ex}\\
        & 1,000,000 & $\underline{62.6}$ & $\underline{62.8}$ & $93.0$ & $\underline{82.7}$ & $\underline{77.8}$ & $\underline{72.8}$ \rule{0pt}{2.3ex}\\
        & 2,000,000 & $61.9$ & $62.7$ & $\underline{93.2}$ & $82.4$ & $77.6$ & $72.6$ \rule{0pt}{2.3ex}\\
        & 4,791,771 & $\mathbf{63.4}$ & $\mathbf{63.1}$ & $\mathbf{93.6}$ & $\mathbf{83.6}$ & $\mathbf{78.5}$ & $\mathbf{73.4}$ \rule{0pt}{2.3ex}\\
        \bottomrule
    \end{tabular}}
    \caption{Impact of increasing the quantity of single-camera data. The model is trained on MSMT17 as multi-camera data in both Mt~(first row) and MSt (other rows) setups.}
    \label{tab:synth-quant}
\end{table}

\label{sec:ablation-matching-loss}
\vspace{2pt}\noindent\textbf{Effect of Matching Loss Design.} We study the impact of different loss functions for image-text matching in \cref{tab:ablation-study-itm-losses}. Replacing the CLIP loss with its soft version (Soft CLIP), which accounts for multiple positive image-text pairs, improves performance. This confirms that soft alignment is beneficial when several captions share the same identity within a mini-batch. However, our proposed Identity-aware Matching Loss~($\mathcal{L}_{im}$) learns the full distribution over all positive and negative instances, outperforming Soft CLIP, which still relies on pairwise similarity. This enables richer and more flexible supervision. Finally, adding the Structure-preserving Loss ($\mathcal{L}_{sp}$) further improves the results by enforcing intra-class consistency between image embeddings in the shared embedding space.

\vspace{2pt}\noindent\textbf{Impact of Increasing the Quantity of Single-Camera Data.} We assess how gradually increasing single-camera textual data affects Re-ID performance~(\cref{tab:synth-quant}). By creating cumulative subsets from the SYNTH-PEDES, where each larger subset contains all samples from the smaller ones, we isolate the impact of additional textual supervision without distributional shifts. The analysis begins with 32,621 samples, equal to the MSMT17 dataset, allowing for direct evaluation of the impact of textual data quantity during training. As the number of captions grows, performance improves consistently, showing a positive correlation between caption volume and Re-ID efficiency. 

\vspace{2pt}\noindent\textbf{Conclusions.} Summing up, the performance gains stem from three factors that distinguish ReText from closely related works: semantic enrichment from textual supervision is evidenced by ~\cref{tab:proof-of-concept-overall} (rows 1--2)\footnote{Textual supervision inherently introduces the use of text-image alignment losses.}, the contribution of the proposed objectives is shown in ~\cref{tab:ablation-study-overall} and ~\cref{tab:ablation-study-itm-losses}, and the data scale effect is isolated in ~\cref{tab:synth-quant}.
\section{Conclusion}
\label{sec:conclusion}

In this paper, we propose ReText, a novel method for generalizable image-based person Re-ID trained on a mixture of multi-camera and single-camera data with textual descriptions. Unlike prior work that ignores single-camera data or underuses language, ReText effectively combines both to learn discriminative, domain-invariant representations. We also fine-tuned VLM to generate texts for multi-camera datasets and assessed the impact of the textual descriptions without additional data. Future work could focus on reducing noise in generated descriptions and enhancing their granularity and consistency to improve Re-ID performance. We believe our work will serve as a basis for future research on generalized person Re-ID.
{
    \small
    \bibliographystyle{ieeenat_fullname}
    \bibliography{main}
}

\clearpage
\setcounter{page}{1}
\maketitlesupplementary

\section{Some Additional Implementation Details}
\label{sec:additional-implementation-details}

\subsection{Mini-batch Composition}

Each mini-batch in ReText (MSt) consists of a mixture of multi-camera and single-camera data:
\begin{itemize}
    \item For the multi-camera portion, we use the well-known $PK$ sampling strategy~\cite{hermans2017defense}, following ReMix \cite{mamedov2025remix}: $P_{m} = 8$ person identities are randomly selected, and for each identity, $K_{m} = 4$ images from different cameras are sampled, resulting in $N_{m} = P_{m} \times K_{m} = 8 \times 4 = 32$ images.
    \item For the single-camera portion, we similarly apply the $PK$ sampler with $P_{s} = 32$ and $K_{s} = 2$, but without enforcing camera diversity, since each identity appears in only one camera. This yields an extra $N_{s} = 64$ images per batch.
\end{itemize}
In total, each mini-batch consists of $96$ images: $32$ from multi-camera data and $64$ from single-camera data. A detailed analysis of the sampling parameters for single-camera data is provided in the following sections (\cref{sec:k-s-analysis}). 

\subsection{Decoder Architecture}
The decoder, used for the image reconstruction task~(\cref{fig:method-scheme}), is responsible for predicting missing image patches based on visible image patches and textual tokens. It consists of four repeated transformer-based decoder blocks, each comprising: a self-attention layer; a cross-attention layer, enabling information flow from the text tokens; a feed-forward network. Finally, all output tokens are projected into pixel space via a linear layer, which reconstructs the masked patches of the image.

Although ReText training involves a text encoder and a decoder alongside the main vision encoder, it remains efficient, requiring $\sim$3 hours~(9 min/epoch) on MSMT17 without extra data and $\sim$8 hours~(24 min/epoch) with single-camera data until convergence is reached by the $20$th epoch.

\section{Using Single-camera Data in Re-ID Losses}
\label{suppl:single-camera-reid-losses}

In ReText, we adopt the Instance ($\mathcal{L}_{ins}$), Augmentation~($\mathcal{L}_{aug}$), Centroids~($\mathcal{L}_{cen}$), and Camera Centroids ($\mathcal{L}_{cc}$) losses from ReMix and DynaMix for the Re-ID task. Unlike ReMix and DynaMix, which apply these loss functions~(except for $\mathcal{L}_{cc}$) to both multi-camera and single-camera data, ReText restricts their use to multi-camera data only. As justified in the main paper, Re-ID supervision is most effective when cross-view variation is present, which is inherently lacking in single-camera data. The latter is stylistically diverse but considerably simpler and less informative for direct cross-view identity classification.

\begin{table}[t]
    \centering
    \fontsize{8pt}{11pt}\selectfont
    \tabcolsep=2.5pt
    \scalebox{0.85}{
    \begin{tabular}{l|c|cc|cc}
        \toprule
        \multirow{2}{*}{Configuration} & \multirow{2}{*}{Text} & \multicolumn{2}{c|}{CUHK03-NP} & \multicolumn{2}{c}{Market-1501} \\
        & & $Rank_1$ & $mAP$ & $Rank_1$ & $mAP$ \\
        \midrule
        in all Re-ID losses & \ding{55} & $49.9$ & $51.3$ & $92.4$ & $78.2$ \\
        in all Re-ID losses & \ding{51} & $63.4$ & $62.9$ & $93.3$ & $82.8$ \\
        in no Re-ID losses & \ding{51} & $63.4$ & $\mathbf{63.1}$ & $\mathbf{93.6}$ & $\mathbf{83.6}$ \\
        only in $\mathcal{L}_{ins}$ & \ding{51} & $63.1$ & $62.9$ & $\mathbf{93.6}$ & $\mathbf{83.6}$ \\
        only in $\mathcal{L}_{aug}$ & \ding{51} & $62.6$ & $62.3$ & $93.5$ & $83.3$ \\
        only in $\mathcal{L}_{cen}$ & \ding{51} & $\mathbf{64.2}$ & $62.8$ & $93.3$ & $82.3$ \\
        \bottomrule
    \end{tabular}}
    \caption{Impact of using single-camera data in different Re-ID loss functions. The model is trained on MSMT17.}
    \label{tab:re-id-losses}
\end{table}

To verify this hypothesis, we conduct a series of experiments in which Re-ID loss functions are selectively applied to single-camera data (\cref{tab:re-id-losses}). 
In most cases, adding Re-ID supervision to single-camera data leads to a degradation in cross-domain performance, confirming our design choice.

The first 
three rows of the table are especially instructive. In the first case, the model is trained without text supervision, and Re-ID losses are applied to both multi-camera and single-camera data. In the second case, we introduce textual supervision 
without any other changes. In the third case, we also utilize textual supervision but apply Re-ID losses only to multi-camera data. The significant performance gain observed in the second setting highlights a key insight: while single-camera data is less complex for person Re-ID, pairing it with descriptive captions unlocks rich semantic cues that significantly enhance domain generalization. The best results are achieved when single-camera data is excluded from Re-ID losses and instead used only for auxiliary tasks such as image-text matching and reconstruction.

\section{Parameter Analysis}
\label{sec:parameter-analysis}

\subsection{Effect of Same-identity Instances in a Mini-batch}
\label{sec:k-s-analysis}

To determine the optimal number of same-identity instances from single-camera data in a mini-batch, we analyze different image-text matching loss functions under varying values of $K_{s}$ (\cref{tab:k-s-analysis}). As shown in the results, performance 
\begin{table}
    \centering
    \fontsize{8pt}{11pt}\selectfont
    \tabcolsep=2.5pt
    \scalebox{0.8}{
    \begin{tabular}{l|c|cc}
        \toprule
        Loss & $K_{s}$ & $Rank_1$ & $mAP$ \\
        \midrule
        \multirow{3}{*}{CLIP loss} & $1$ & $59.8$ & $60.3$ \\
        & $2$ & $59.6$ & $60.1$ \\
        & $4$ & $59.5$ & $59.2$ \\
        \midrule
        \multirow{3}{*}{Soft CLIP loss} & $1$ & $59.8$ & $60.3$ \\
        & $2$ & $60.0$ & $60.5$ \\
        & $4$ & $59.0$ & $59.5$ \\
        \midrule
        \multirow{3}{*}{$\mathcal{L}_{im}$ (ours)} & $1$ & $60.6$ & $60.6$ \\
        & $2$ & $\mathbf{61.4}$ & $\mathbf{61.3}$ \\
        & $4$ & $60.4$ & $60.0$ \\
        \bottomrule
    \end{tabular}}
    \caption{Analysis of the number of same-identity instances~($K_{s}$) from single-camera data in a mini-batch for different image-text matching loss functions. The model is trained on MSMT17 and tested on CUHK03-NP.}
    \label{tab:k-s-analysis}
\end{table}
under the standard CLIP contrastive loss decreases as $K_{s}$ increases. This is expected, as CLIP performs strict one-to-one matching and cannot effectively handle multiple positive image-text pairs per identity in a mini-batch. In contrast, Soft CLIP loss --- which accounts for multiple positive image-text pairs --- shows improved performance when $K_{s} = 2$, confirming that soft alignment is beneficial in this setting. The best performance is achieved using our Identity-aware Matching Loss ($\mathcal{L}_{im}$) with $K_{s} = 2$. This highlights not only its robustness for image-text alignment but also its flexibility in handling multiple positive image-text pairs per identity in a mini-batch.

\textbf{Note:} in these experiments, the number of single-camera images in a mini-batch ($N_{s}$) was fixed to $32$, and we used $\alpha = 0.5$ in \cref{eq:q-in-im-loss}. Therefore, selecting $K_{s} = 2$ implies that $N_{pos} = 1$ in \cref{eq:q-in-im-loss}.

\begin{table}[t]
\centering
\scalebox{0.9}{
\begin{subtable}[t]{0.32\linewidth}
\vspace{0pt}
\centering
\fontsize{8pt}{11pt}\selectfont
\tabcolsep=2.5pt
\begin{tabular}{c|cc}
    \toprule
    $N_{s}$ & $Rank_1$ & $mAP$ \rule{0pt}{2.3ex}\rule[-0.9ex]{0pt}{0pt}\\
    \midrule
    $32$ & $61.4$ & $61.3$ \rule{0pt}{2.3ex}\\
    $64$ & $61.6$ & $\mathbf{61.9}$ \rule{0pt}{2.3ex}\\
    $128$ & $\mathbf{61.9}$ & $61.8$ \rule{0pt}{2.1ex}\\
    \bottomrule
\end{tabular}
\caption{}
\label{tab:number-of-single-camera-images-analysis}
\end{subtable}
\hfill
\begin{subtable}[t]{0.32\linewidth}
\vspace{0pt}
\centering
\fontsize{8pt}{11pt}\selectfont
\tabcolsep=2.5pt
\begin{tabular}{c|cc}
    \toprule
    $\alpha$ & $Rank_1$ & $mAP$ \\
    \midrule
    $0.5$ & $61.6$ & $61.9$ \\
    $0.6$ & $\mathbf{62.2}$ & $\mathbf{62.3}$ \\
    $0.7$ & $61.9$ & $62.0$ \\
    \bottomrule
\end{tabular}
\caption{}
\label{tab:alpha-analysis}
\end{subtable}
\hfill
\begin{subtable}[t]{0.32\linewidth}
\vspace{0pt}
\centering
\fontsize{8pt}{11pt}\selectfont
\tabcolsep=2.5pt
\begin{tabular}{c|cc}
    \toprule
    $\tau$ & $Rank_1$ & $mAP$ \\
    \midrule
    $0.07$ & $62.6$ & $62.5$ \\
    $0.1$ & $\mathbf{62.9}$ & $\mathbf{62.7}$ \\
    $0.15$ & $61.9$ & $62.3$ \\
    \bottomrule
\end{tabular}
\caption{}
\label{tab:tau-analysis}
\end{subtable}}
\caption{a) Analysis of the number of single-camera images in a mini-batch ($N_{s}$). b) Analysis of the $\alpha$ parameter in \cref{eq:q-in-im-loss} for the Identity-aware Matching Loss. c) Analysis of the temperature parameter $\tau$ in \cref{eq:sp-loss} for the Structure-preserving Loss. The model is trained on MSMT17 and tested on CUHK03-NP.}
\end{table}

\subsection{Number of Single-camera Images in a Mini-batch}
\label{sec:number-of-single-camera-images-analysis}
We study how the number of single-camera images in a mini-batch ($N_{s}$) affects cross-domain Re-ID performance. As shown in \cref{tab:number-of-single-camera-images-analysis}, increasing $N_{s}$ from $32$ to $64$ and $128$ leads to gradual improvements. This trend suggests that more single-camera data --- when paired with descriptive captions~--- provides stronger semantic supervision and greater visual diversity, thereby improving generalization. However, the gains saturate beyond $64$ images, indicating a trade-off between diversity and computational efficiency.

\textbf{Note:} in these experiments, the number of same-identity instances~($K_{s}$) from single-camera data in a mini-batch was fixed to $2$, and we used $\alpha = 0.5$ in \cref{eq:q-in-im-loss}.

\subsection{Controlling Positive Image-Text Pairs Weighting with $\alpha$}


We analyze the effect of the $\alpha$ parameter in \cref{eq:q-in-im-loss}, which controls the relative weight of the main image-text pair versus other positive pairs in the Identity-aware Matching Loss. As shown in \cref{tab:alpha-analysis}, the best performance is achieved with $\alpha = 0.6$, indicating that slightly prioritizing the anchor pair while still leveraging additional positives yields the most robust alignment. Overweighting additional captions (i.e., setting lower $\alpha$) can be risky: although the images refer to the same identity, the corresponding textual descriptions may differ due to variations in pose, visibility, or context. For instance, sneakers might be visible in one image but occluded in another, resulting in different caption content. Nevertheless, these additional captions provide valuable complementary semantic cues. Our loss formulation strikes a balance --- leveraging semantic diversity across views while preserving stability in the matching process.

\textbf{Note:} in these experiments, the number of single-camera images in a mini-batch ($N_{s}$) was fixed to $64$, and we used $K_{s} = 2$.

\subsection{Effect of The Temperature Parameter $\tau$ in The Structure-preserving Loss}

We study the influence of the temperature parameter $\tau$ in \cref{eq:sp-loss}, which controls the sharpness of similarity distributions in the Structure-preserving Loss. As shown in \cref{tab:tau-analysis}, setting $\tau = 0.1$ yields the best performance. This value provides a good trade-off between emphasizing hard positives and maintaining stable gradients.

Too small a value (e.g., $\tau = 0.07$) makes the loss overly sensitive to small similarity differences, while larger values (e.g., $\tau \geq 0.15$) overly smooth the similarity scores, weakening the structure-preserving effect. These results confirm that fine-tuning $\tau$ is crucial for enforcing intra-class structure without destabilizing training.

\textbf{Note:} in these experiments, the number of single-camera images in a mini-batch ($N_{s}$) was fixed to $64$, $K_{s} = 2$, and we used $\alpha = 0.6$.

\begin{figure*}[t]
    \centering
    \includegraphics[width=0.9\linewidth]{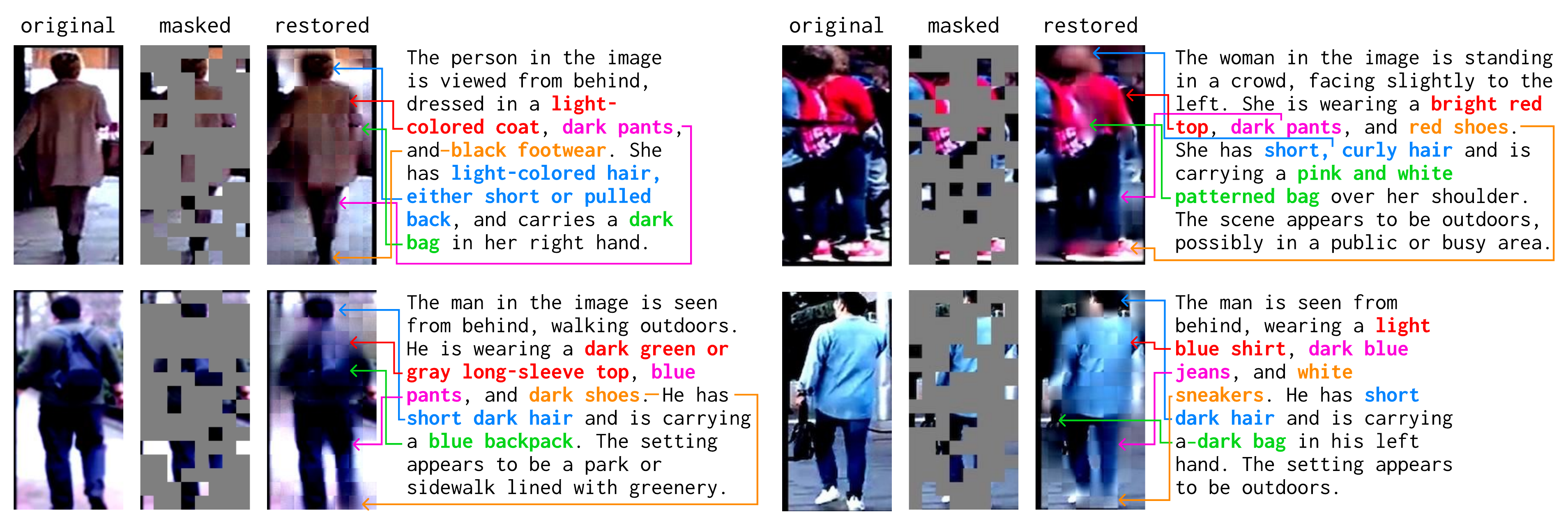}
    \caption{Examples of image reconstruction in ReText. Each example shows the original image~(left), masked input (middle), and the reconstructed output (right), alongside the corresponding caption. The model successfully recovers key visual details, highlighting the effectiveness of image-text training.}
    \label{fig:image-reconstruction}
\end{figure*}

\section{Image Reconstruction}
\label{sec:image-reconstruction}

In our experiments, we mask $75\%$ of image patches. As illustrated in \cref{fig:image-reconstruction}, ReText successfully reconstructs the human silhouette and key attributes. These examples also highlight the influence of text supervision: the model is able to recover attributes that are entirely masked out in the image, guided by their presence in the accompanying description. It is important to note that reconstruction serves as an auxiliary task --- our goal is not to achieve photorealistic quality but to improve generalization. As shown in \cref{tab:ablation-study-overall}, the image reconstruction task enhances cross-domain Re-ID performance by helping ReText learn robust representations under partial visual information and domain shift.

\section{Some Additional Experiments Results}
\label{sec:protocols}
In addition to the comparison using the standard Protocol 1 (\cref{sec:protocol1}), we also provide experimental results by two other protocols --- Protocol 2 and Protocol~3, as they are frequently cited in the literature for general comparison of our method with others. It both utilizes several multi-camera datasets for training and evaluates on a hold-out target dataset, while Protocol 3 also includes both train and test splits of the source datasets for training.

\begin{table}[t]
    \centering
    \fontsize{8pt}{11pt}\selectfont
    \tabcolsep=2.5pt
    \scalebox{0.72}{
    \begin{tabular}{@{\hspace{1pt}}l@{\hskip 6pt}l|cc|cc|cc|cc}
        \toprule
        & \multirow{2}{*}{Method} & \multicolumn{2}{c|}{M+MS+CS $\rightarrow$ C3} & \multicolumn{2}{c|}{M+CS+C3 $\rightarrow$ MS} & \multicolumn{2}{c|}{MS+CS+C3 $\rightarrow$ M} & \multicolumn{2}{c}{Average}\\
        & & $Rank_1$ & $mAP$ & $Rank_1$ & $mAP$ & $Rank_1$ & $mAP$ & $Rank_1$ & $mAP$\rule{0pt}{2.3ex}\rule[-0.5ex]{0pt}{0pt}\\
        \midrule
        \multirow{10}{*}{\rotatebox{90}{Protocol 2}} & META \cite{xu2021meta} & $35.1$ & $36.3$ & $49.9$ & $22.5$ & $86.1$ & $67.5$ & $57.0$ & $42.1$ \rule{0pt}{2.3ex}\\
        & ACL \cite{zhang2022adaptive} & $41.8$ & $41.2$ & $45.9$ & $20.4$ & $89.3$ & $74.3$ & $59.0$ & $45.3$ \rule{0pt}{2.3ex}\\
        & CLIP-ReID \cite{li2023clip} & $41.9$ & $42.1$ & $53.1$ & $26.6$ & $84.4$ & $68.8$ & $59.8$ & $45.8$ \rule{0pt}{2.3ex}\\
        & ReFID \cite{peng2024refid} & $34.8$ & $33.3$ & $39.8$ & $18.3$ & $85.3$ & $67.6$ & $53.3$ & $39.7$ \rule{0pt}{2.3ex}\\
        & GMN \cite{qi2024generalizable} & $42.1$ & $43.2$ & $50.9$ & $24.4$ & $87.1$ & $72.3$ & $60.0$ & $46.6$ \rule{0pt}{2.3ex}\\
        & CLIP-DFGS \cite{zhao2024clip} & $51.1$ & $50.4$ & $59.7$ & $31.5$ & $90.5$ & $79.0$ & $67.1$ & $53.6$ \rule{0pt}{2.3ex}\\
        & BAU \cite{cho2024generalizable} & $43.9$ & $42.8$ & $50.9$ & $24.3$ & $90.4$ & $77.1$ & $61.7$ & $48.1$ \rule{0pt}{2.3ex}\\
        & CLIP-FGDI \cite{zhao2025cilp} & $44.6$ & $44.4$ & $59.4$ & $31.1$ & $\underline{91.3}$ & $79.4$ & $65.1$ & $51.6$ \rule{0pt}{2.3ex}\\
        & ReText (Mt) & $\mathbf{65.7}$ & $\mathbf{66.8}$ & $\underline{61.1}$ & $\underline{36.7}$ & $\mathbf{91.6}$ & $\mathbf{82.4}$ & $\underline{72.8}$ & $\underline{62.0}$ \rule{0pt}{2.1ex}\\
        & ReText (MSt) & $\underline{65.6}$ & $\underline{65.7}$ & $\mathbf{63.4}$ & $\mathbf{39.3}$ & $\underline{91.3}$ & $\underline{82.1}$ & $\mathbf{73.4}$ & $\mathbf{62.4}$ \rule{0pt}{2.1ex}\\
        \midrule
        \multirow{10}{*}{\rotatebox{90}{Protocol 3}} & META \cite{xu2021meta} & $46.2$ & $47.1$ & $52.1$ & $24.4$ & $90.5$ & $76.5$ & $62.9$ & $49.3$ \rule{0pt}{2.3ex}\\
        & ACL \cite{zhang2022adaptive} & $50.1$ & $49.4$ & $47.3$ & $21.7$ & $90.6$ & $76.8$ & $62.7$ & $49.3$ \rule{0pt}{2.3ex}\\
        & CLIP-ReID \cite{li2023clip} & $45.8$ & $44.9$ & $52.6$ & $26.8$ & $83.4$ & $67.5$ & $60.6$ & $46.4$ \rule{0pt}{2.3ex}\\
        & ReFID \cite{peng2024refid} & $44.2$ & $45.5$ & $43.3$ & $20.6$ & $87.9$ & $72.5$ & $58.5$ & $46.2$ \rule{0pt}{2.3ex}\\
        & GMN \cite{qi2024generalizable} & $50.1$ & $49.5$ & $51.0$ & $24.8$ & $89.0$ & $75.9$ & $63.4$ & $50.1$ \rule{0pt}{2.3ex}\\
        & CLIP-DFGS \cite{zhao2024clip} & $51.3$ & $51.6$ & $62.0$ & $33.4$ & $91.6$ & $81.0$ & $68.3$ & $55.3$ \rule{0pt}{2.3ex}\\
        & BAU \cite{cho2024generalizable} & $51.8$ & $50.6$ & $54.3$ & $26.8$ & $91.1$ & $79.5$ & $65.7$ & $52.3$ \rule{0pt}{2.3ex}\\
        & CLIP-FGDI \cite{zhao2025cilp} & $50.1$ & $50.1$ & $61.4$ & $32.9$ & $91.2$ & $79.8$ & $67.6$ & $54.3$ \rule{0pt}{2.3ex}\\
        & ReText (Mt) & $\mathbf{69.3}$ & $\mathbf{70.2}$ & $\underline{62.1}$ & $\underline{37.7}$ & $\mathbf{93.3}$ & $\mathbf{85.2}$ & $\underline{74.9}$ & $\underline{64.4}$ \rule{0pt}{2.1ex}\\
        & ReText (MSt) & $\underline{68.4}$ & $\underline{69.2}$ & $\mathbf{64.6}$ & $\mathbf{40.3}$ & $\underline{93.2}$ & $\underline{84.1}$ & $\mathbf{75.4}$ & $\mathbf{64.5}$ \rule{0pt}{2.1ex}\\
        \bottomrule
    \end{tabular}}
    \caption{Comparison of ReText with other state-of-the-art methods according to Protocols 2 and 3. In this table, C3 is CUHK03-NP, MS is MSMT17, M is Market-1501, and CS is CUHK-SYSU. \textit{ReMix and DynaMix are excluded from this comparison because its authors did not evaluate the method using these protocols.}}
    \label{tab:comparison-multi}
\end{table}

\begin{table}[t]
    \centering
    \fontsize{8pt}{11pt}\selectfont
    \tabcolsep=2.5pt
    \scalebox{0.83}{
    \begin{tabular}{@{\hspace{7pt}}l|c|cc|cc|cc}
        \toprule
        \multirow{2}{*}{Method} & \multirow{2}{*}{\rotatebox{90}{Tr. D.}} & \multicolumn{2}{c|}{CUHK03-NP} & \multicolumn{2}{c|}{Market-1501} & \multicolumn{2}{c}{Average}\\
        & & $Rank_1$ & $mAP$ & $Rank_1$ & $mAP$ & $Rank_1$ & $mAP$ \rule[-0.5ex]{0pt}{0pt}\\
        \midrule
        QAConv \cite{liao2020interpretable} & \multirow{9}{*}{\rotatebox{90}{MSMT17-merged}} & $25.3$ & $22.6$ & $72.6$ & $43.1$ & $49.0$ & $32.9$ \rule{0pt}{2.3ex}\\
        TransMatcher \cite{liao2021transmatcher} & & $31.9$ & $30.7$ & $82.6$ & $58.4$ & $57.3$ & $44.6$ \rule{0pt}{2.3ex}\\
        QAConv-GS \cite{liao2022graph} & & $27.6$ & $28.0$ & $82.4$ & $56.9$ & $55.0$ & $42.5$ \rule{0pt}{2.3ex}\\
        PAT \cite{ni2023part} & & $27.4$ & $28.7$ & $72.8$ & $48.6$ & $50.1$ & $38.7$ \rule{0pt}{2.3ex}\\
        CLIP-DFGS \cite{zhao2024clip} & & $37.1$ & $25.7$ & $80.9$ & $55.2$ & $59.0$ & $40.5$ \rule{0pt}{2.3ex}\\
        ReMix \cite{mamedov2025remix} & & $37.7$ & $37.2$ & $84.0$ & $61.0$ & $60.9$ & $49.1$ \rule{0pt}{2.3ex}\\
        DynaMix \cite{mamedov2025dynamix} & & $60.9$ & $61.0$ & $92.6$ & $80.8$ & $76.8$ & $70.9$ \rule{0pt}{2.3ex}\\
        ReText (Mt) & & $\underline{66.0}$ & $\underline{66.2}$ & $\mathbf{94.2}$ & $\mathbf{84.8}$ & $\mathbf{80.1}$ & $\underline{75.5}$ \rule{0pt}{2.1ex}\\
        ReText (MSt) & & $\mathbf{66.1}$ & $\mathbf{66.8}$ & $\underline{93.9}$ & $\underline{84.7}$ & $\underline{80.0}$ & $\mathbf{75.8}$ \rule{0pt}{2.1ex}\\
        \bottomrule
    \end{tabular}}
    \caption{Comparison of ReText with other state-of-the-art methods according to Protocol 1 using the MSMT17-merged training dataset.}
    \label{tab:comparison-single2}
\end{table}

When trained on several multi-camera datasets, ReText demonstrates strong and consistent performance across all target domains~(\cref{tab:comparison-multi}), highlighting its ability to benefit not only from stylistic diversity in single-camera data and semantic supervision provided by textual descriptions, but also from domain variation in multi-camera datasets. Notably, ReText significantly outperforms CLIP-ReID \cite{li2023clip}, CLIP-DFGS \cite{zhao2024clip}, and CLIP-FGDI \cite{zhao2025cilp}, which rely on learnable text tokens instead of descriptive captions associated with images. This comparison confirms our hypothesis that the semantic richness and generalization capability of natural language remain underutilized in current image-based person Re-ID approaches. By directly leveraging textual descriptions, ReText effectively incorporates semantic cues that substantially boost cross-domain generalization. The Mt version of ReText surpasses the MSt version in several cases, indicating that the number of samples in the combination of multi-camera data is sufficient.

\begin{table}[t]
    \centering
    \fontsize{8pt}{11pt}\selectfont
    \tabcolsep=2.5pt
    \scalebox{0.77}{
    \begin{tabular}{@{\hspace{7pt}}l|cc|cc|cc}
        \toprule
        \multirow{2}{*}{LLaVA} & \multicolumn{2}{c|}{CUHK03-NP} & \multicolumn{2}{c|}{Market-1501} & \multicolumn{2}{c}{Average}\\
        & $Rank_1$ & $mAP$ & $Rank_1$ & $mAP$ & $Rank_1$ & $mAP$ \rule[-0.5ex]{0pt}{0pt}\\
        \midrule
        HAM~\cite{jiang2025modeling} & $\underline{60.0}$ & $\underline{61.1}$ & $\mathbf{93.6}$ & $\underline{81.4}$ & $\underline{76.8}$ & $\underline{71.2}$ \rule{0pt}{2.3ex}\\
        Ours (CIR) & $59.7$ & $60.7$ & $\underline{93.5}$ & $\underline{81.4}$ & $76.6$ & $71.1$ \rule{0pt}{2.3ex}\\
        Ours (UFine3C) & $\mathbf{61.9}$ & $\mathbf{61.8}$ & $93.3$ & $\underline{81.4}$ & $\mathbf{77.6}$ & $\mathbf{71.6}$ \rule{0pt}{2.3ex}\\
        Ours (UFine3C$^\dagger$) & $59.7$ & $60.6$ & $\mathbf{93.6}$ & $\mathbf{81.5}$ & $76.7$ & $71.0$ \rule{0pt}{2.3ex}\\
        \bottomrule
    \end{tabular}}
    \caption{Comparison of LLaVA models fine-tuned on different datasets. The CIR is the combination of CUHK-PEDES, ICFG-PEDES and RSTPReid datasets. UFine3C$^\dagger$ denotes LLaVA fine-tuned on UFine3C, excluding images overlapping with the multi-camera datasets. The model is trained on MSMT17.}
    \label{tab:comparison-data}
\end{table}

\begin{table}[tb]
    \centering
    \scalebox{0.7}{
    \begin{tabular*}{0.55\textwidth}{@{\hspace{8pt}}@{\extracolsep{\fill}}lccc}
        \toprule
        Dataset & \#Texts & \#Texts w. reps. & \#Too Long Texts \\
        \midrule
        CUHK03-NP & $14,097$ & $1$ & $0$ \\
        Market-1501 & $32,217$ & $15$ & $4$ \\
        MSMT17 & $126,441$ & $127$ & $16$ \\
        CUHK-SYSU & $34,574$ & $174$ & $5$ \\
        RandPerson & $132,145$ & $350$ & $53$ \\
        \bottomrule
    \end{tabular*}}
    \caption{Number of generated captions per dataset, and counts of captions containing repeated phrases (\#Texts w. reps.) and exceeding the length threshold (\#Too Long Texts).}
    \label{tab:filtering}
\end{table}

\begin{figure}[t]
    \centering
    \includegraphics[width=0.906\linewidth]{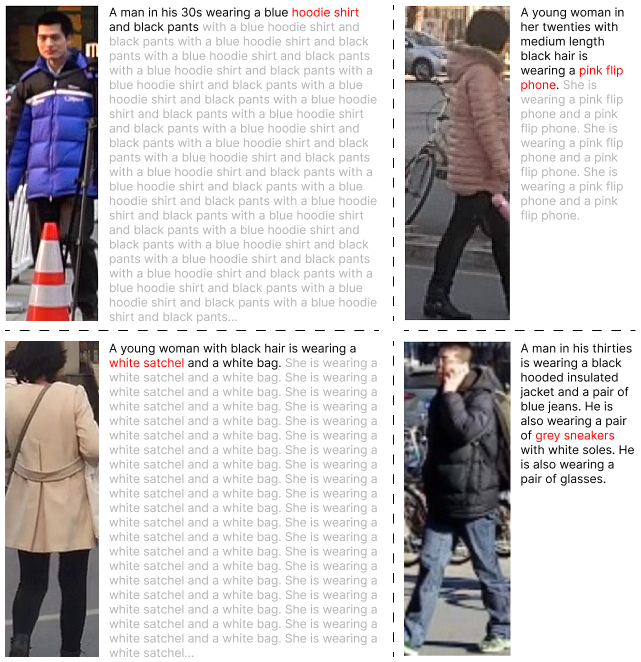}
    \caption{
    Examples of defective generated captions. Hallucinated attributes that are not supported by the image are shown in red text, while repeated phrases produced by degenerate looping in the generation process are shown in gray text.}
    \label{fig:repetitions}
\end{figure}

\noindent\textbf{An Additional Comparison According to Protocol 1.} We present experiments with the MSMT17-merged dataset, which combines training and testing parts of MSMT17, to demonstrate the effect of supplementary multi-camera samples on performance. As shown in \cref{tab:comparison-single2}, ReText surpasses other methods in a generalizable person Re-ID setup. Notably, ReText (Mt) performs more similarly to ReText (MSt) when trained on merged MSMT17 samples than when trained on MSMT17 only samples. It represents the direct dependence of efficiency and quantity of multi-camera data samples paired with texts. Also, the performance gap between training on MSMT17 (\cref{tab:comparison-single}) and MSMT17-merged~(\cref{tab:comparison-single2}) is smaller for ReText compared to methods like ReMix and DynaMix, indicating that ReText relies less on the scale of multi-camera data and benefits more from the semantic richness of textual supervision and the stylistic diversity of single-camera data.

\section{Texts Generation for Image-Based Re-ID Datasets}
\subsection{Search for Suitable Datasets Used in LLaVA Fine-tuning}
\label{sec:data-for-llava}

We search for an optimal dataset combination to fine-tune LLaVA and subsequently generate high-quality descriptions for multi-camera data. First, we fine-tune LLaVa on the CUHK-PEDES~\cite{li2017person} and ICFG-PEDES~\cite{ding2021semantically} datasets using style prompts, as presented in~\cite{jiang2025modeling} (HAM in \cref{tab:comparison-data}). Further, we fine-tune another LLaVA on a mix of common text-based Re-ID datasets --- CUHK-PEDES, ICFG-PEDES, RSTPReid~\cite{zhu2021dssl} (CIR), but it leads to degraded description quality and weaker downstream performance. Finally, the UFine3C~\cite{zuo2024ufinebench} dataset, which is based on three datasets with different granularities --- CUHK-PEDES, ICFG-PEDES, and UFine6926~\cite{zuo2024ufinebench} --- is used for LLaVA fine-tuning and yields superior results compared to the two previously described dataset choices.

Since both CUHK-PEDES and ICFG-PEDES are built from CUHK03, Market-1501, and MSMT17 images, there is an overlap between the UFine3C dataset used for LLaVA and the evaluation sets in the target domain of ReText. We found that UFine3C contains $<1\%$ of the Market-1501 test images, $<2\%$ of the MSMT17 test images, and none from the CUHK03 test set. To isolate the impact of this intersection, we retrain LLaVA with UFine3C, excluding $1,556$ overlapping images, resulting in $5,890$ images instead of $7,446$. Then, we generate new captions for MSMT17 and retrain ReText. As a result, CUHK03 mAP decreased from $61.8$ to $60.6$, while Market-1501 mAP increased from $81.4$ to $81.5$ (\cref{tab:comparison-data}), indicating that the reported gains are not due to this overlap.

Since there is no initial intersection between the UFine3C dataset and the CUHK03 test set, we associate a large drop in performance on it with the reduced LLaVA fine-tuning set. 
Conceptually, the risk of such overlap is limited, since inference is image-only and the VLM is fine-tuned for captioning rather than identity classification. So, any overlap can at most affect caption phrasing or calibration --- it cannot directly leak identity-discriminative signal into the model. Therefore, in all other experiments, where multi-camera datasets are paired with texts, captions are generated by LLaVA fine-tuned on the full UFine3C dataset.

\begin{figure*}[t]
    \centering
    \includegraphics[width=1.0\linewidth]{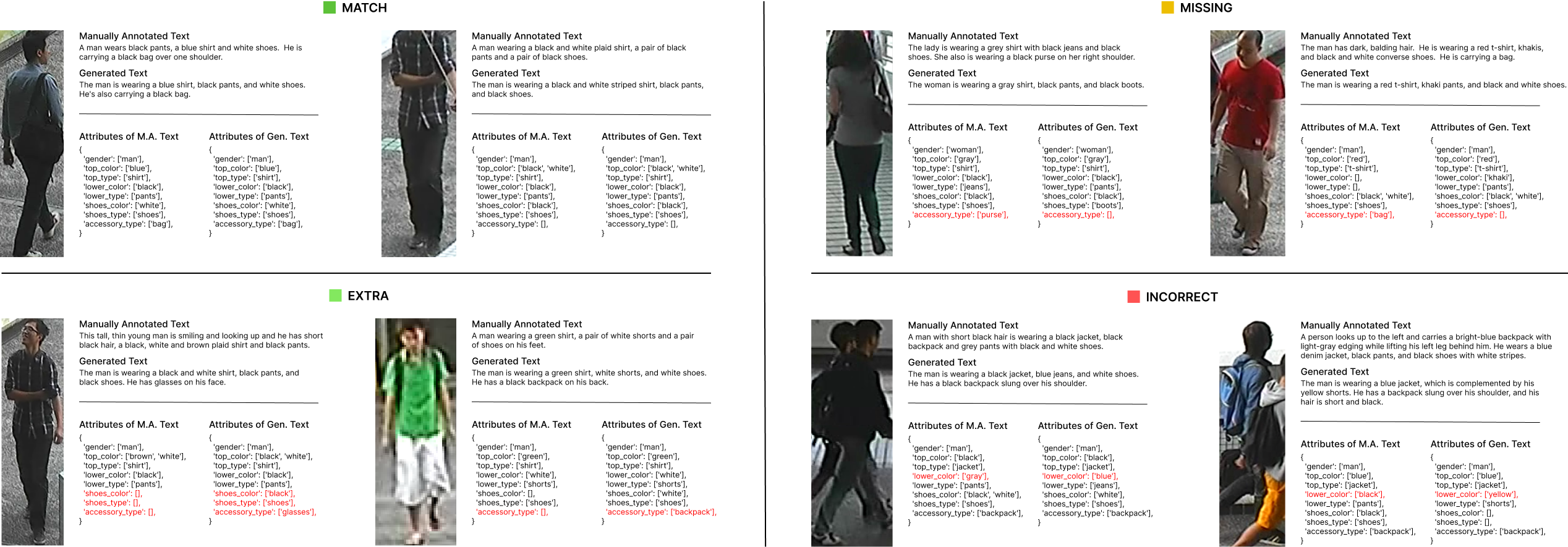}
    \caption{Qualitative examples of the attribute-matching categories from Figure~\ref{fig:pies}, showing manually annotated (M.A.) text, generated~(Gen.) text, and their corresponding extracted attribute sets. The fields that differ between M.A. and Gen. texts are highlighted in red.}
    \label{fig:errors}
\end{figure*}

\begin{figure}[t]
    \centering
    \includegraphics[width=1.0\linewidth]{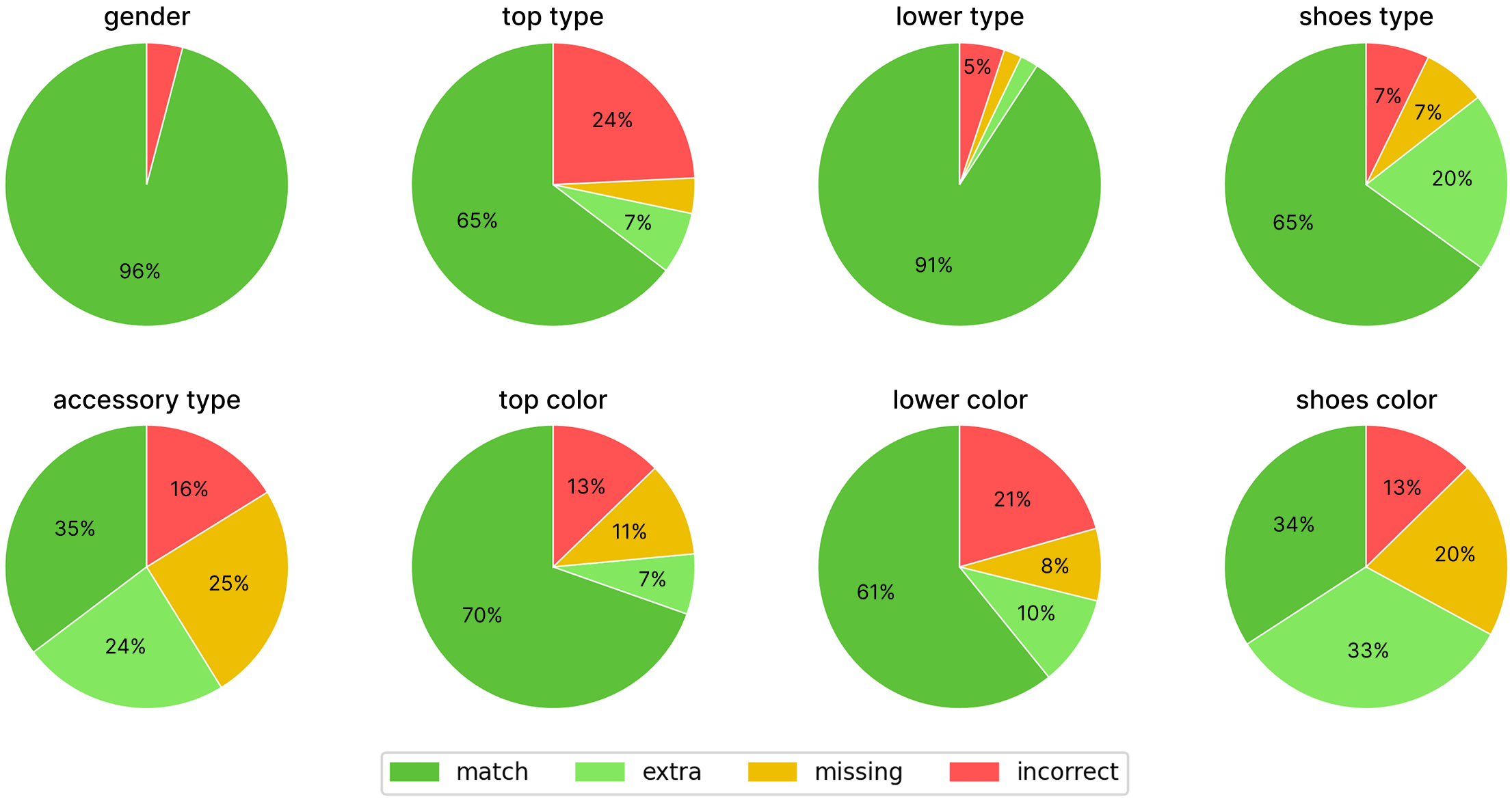}
    \caption{Attribute-level agreement between generated and human-annotated captions. Attributes were manually extracted from both captions and matched. Each chart displays the share of matches (correctly reproduced), extra attributes (noted in the generated text but omitted in the original), missing attributes (present in the original but omitted in the generated), and incorrect values~(conflicting values).}
    \label{fig:pies}
\end{figure}

\subsection{Caption Quality Evaluation}
\label{sec:quality-of-caps}
\vspace{2pt}\noindent\textbf{Caption Filtering.} To assess the reliability of the generation process, we report per-dataset filtering statistics in \cref{tab:filtering}, including the number of captions affected by repetition loops and excessive length. \cref{fig:repetitions} shows representative examples of hallucinated content and repetition loops. 

\vspace{2pt}\noindent\textbf{Caption Quality.} Since the text-to-image ReID dataset, CUHK-PEDES, is constructed using images from, partly, the CUHK03 dataset, there is an opportunity to compare the human-written captions from CUHK-PEDES with the generated captions for the corresponding images in CUHK03. We randomly sample $300$ images from CUHK03, which are not part of UFine3C (see \cref{sec:data-for-llava} for details), and retrieve the corresponding human-written caption from CUHK-PEDES, along with the caption generated by our fine-tuned VLM. For both captions in each pair, we manually extract structured attributes across eight categories: gender, top type, top color, lower type, lower color, shoe type, shoe color, and accessory type. We then match the two attribute sets category by category, treating the human-written caption as ground truth. We assign each comparison to one of four groups: 1) match, when the same value is stated in both captions; 2) missing, when an attribute is mentioned in the human-written caption but omitted from the generated one; 3) extra, when an attribute appears in the generated caption but is absent from the human-written one; and 4) incorrect, when both captions mention the attribute but with conflicting values, indicating a hallucination. \cref{fig:pies} reports the resulting rates per attribute category. Note that an extra attribute is not necessarily an error: since the human-written caption may just omit a detail that is nonetheless visible in the image, such cases cannot be automatically confirmed as correct or incorrect from text alone. Qualitative inspection \cref{fig:errors} shows it is possible that extra attributes are accurate, suggesting that the generated captions frequently capture additional valid details beyond what the reference caption describes.

\begin{figure}
    \centering
    \includegraphics[width=0.8\linewidth]{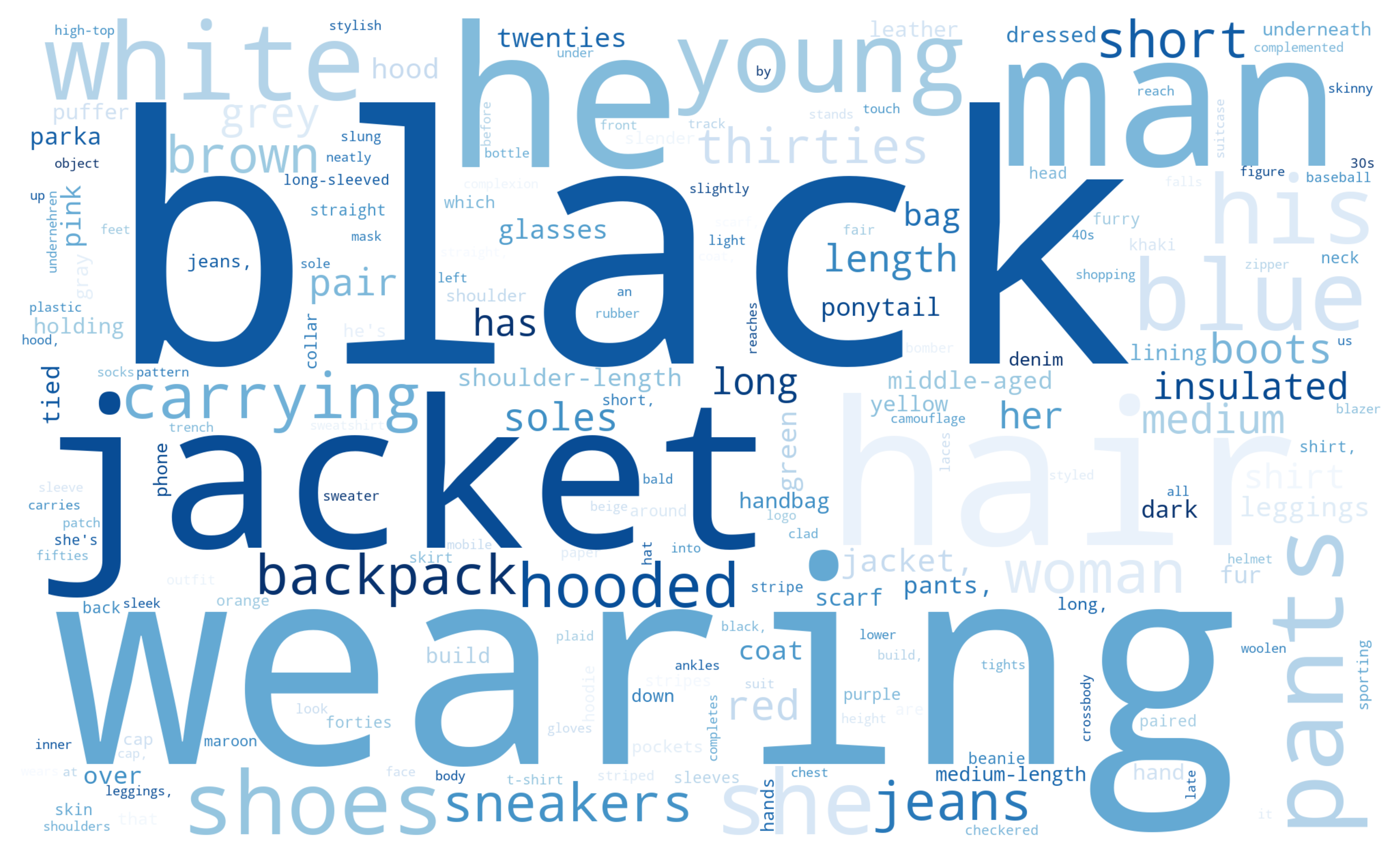}
    \caption{High-frequency words of MSMT17 generated captions.}
    \label{fig:cloud}
\end{figure}

\begin{figure*}[t]
    \centering
    \includegraphics[width=\linewidth]{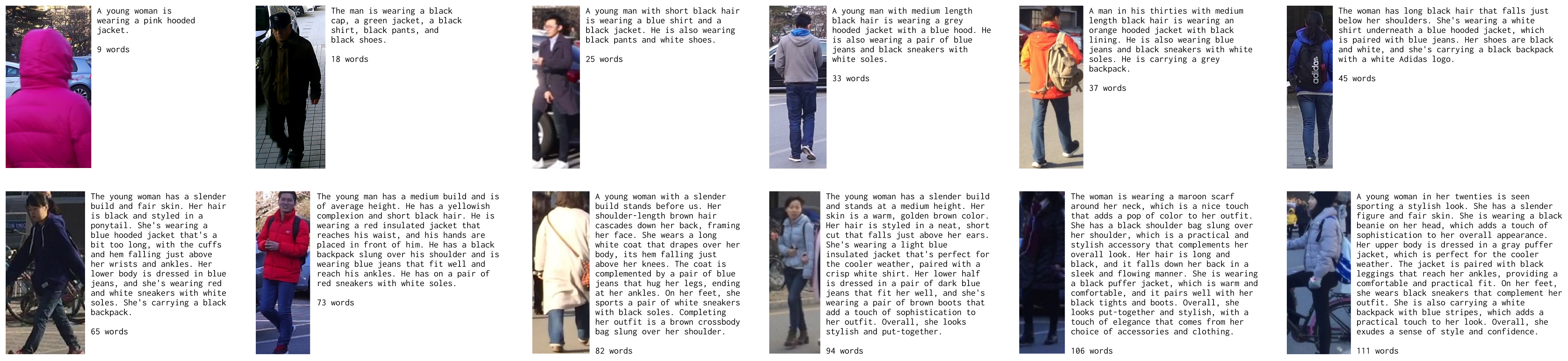}
    \caption{Examples of the textual descriptions generated by LLaVA for the images in the MSMT17 dataset.}
    \label{fig:examples}
\end{figure*}

\begin{figure}[t]
    \centering
    \includegraphics[width=0.9\linewidth]{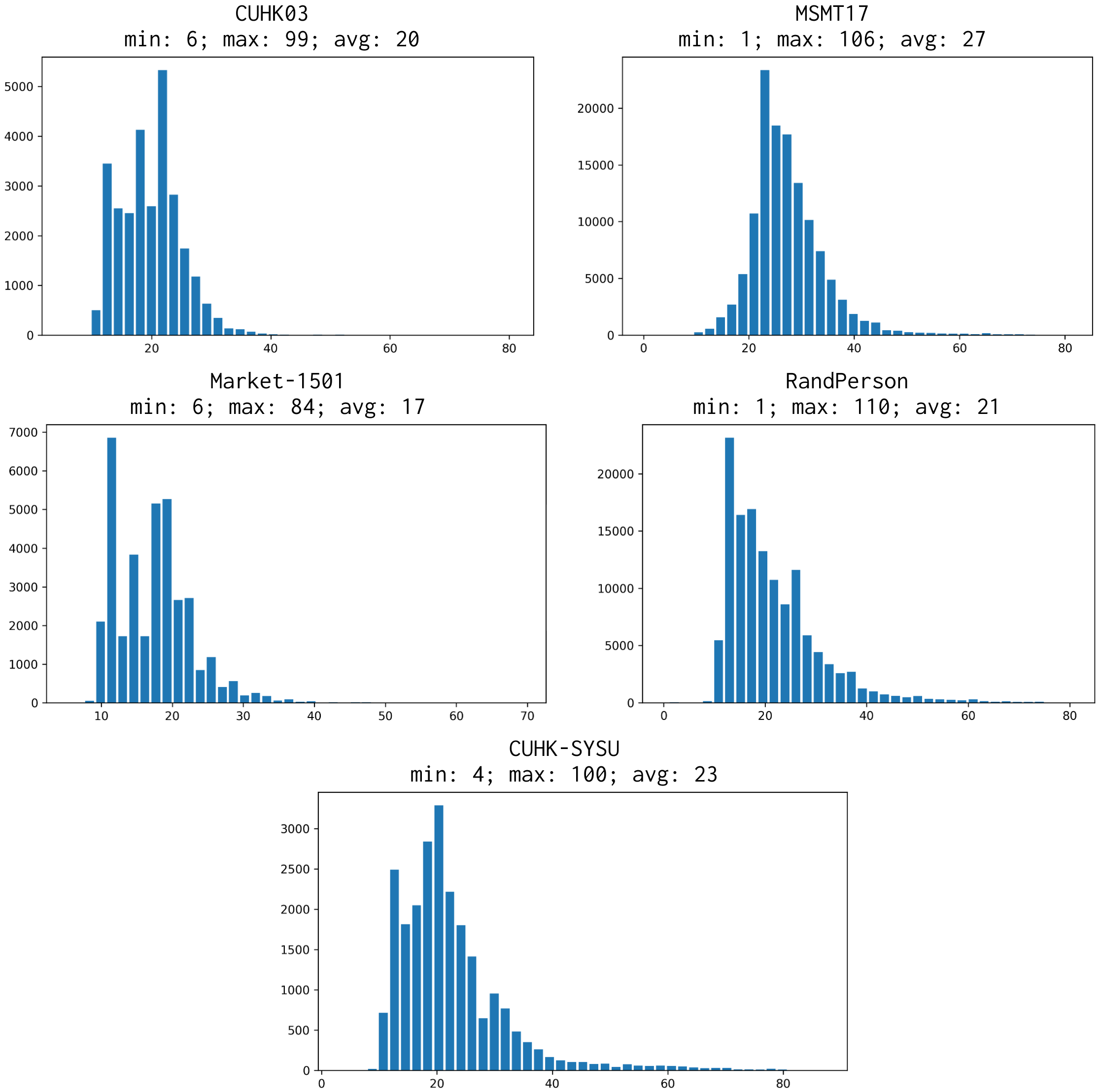}
    \caption{Distribution of the word count in each generated caption for CUHK03, MSMT17, Market-1501, RandPerson, and CUHK-SYSU datasets. Min., max., avg. corresponds to the minimum, maximum, and average number of words.}
    \label{fig:chars}
\end{figure}

\subsection{Some Statistics of Generated Captions}
\label{sec:stat-of-caps}
\cref{fig:cloud} illustrates high-frequency words, while visualization of MSMT17 captions with varying word counts is provided in \cref{fig:examples}. \cref{fig:chars} illustrates the distribution of word counts in the generated captions for widely used multi-camera Re-ID datasets. Across all benchmarks, the distributions exhibit a unimodal shape with a concentration in the $15–30$ word range. Overall, the distributions confirm that the generated captions maintain consistent verbosity across datasets while adapting their length to dataset-specific visual complexity.

\section{Future Research}
Future research could explore improving the granularity and consistency of textual descriptions to further enhance both Re-ID accuracy and reconstruction quality. It could also investigate noise reduction in automatically generated descriptions and assess its impact on quality. Besides, while textual supervision helps mitigate domain shift, it does not eliminate it, as descriptions may still vary with illumination changes, highlighting an important direction for future research.

Moreover, our findings indicate that single-camera data~--- when paired with descriptive text ---~becomes a valuable source of diversity and supervision. However, optimal sampling strategies from such data remain an open question. Exploring better mini-batch composition, instance balancing, or caption-aware sampling may lead to further gains.


\end{document}